\documentclass[12pt,a4paper,oneside,english]{amsart}
\usepackage[T1]{fontenc}
\usepackage[utf8]{inputenc}
\usepackage{babel}
\usepackage{amsbsy}
\usepackage{amstext}
\usepackage{amsthm}
\usepackage{amssymb}
\usepackage{graphicx}
\graphicspath{{Figures/}} 
\usepackage{tikz}
\usetikzlibrary{arrows.meta,fit,backgrounds}

\usepackage[hidelinks]{hyperref}

\makeatletter
\numberwithin{equation}{section}
\numberwithin{figure}{section}
\theoremstyle{plain}
\newtheorem{thm}{\protect\theoremname}
\theoremstyle{definition}
\newtheorem{defn}[thm]{\protect\definitionname}
\theoremstyle{remark}
\newtheorem{rem}[thm]{\protect\remarkname}
\theoremstyle{plain}

\theoremstyle{plain}
\newtheorem{lem}[thm]{\protect\lemmaname}
\theoremstyle{plain}
\newtheorem{prop}[thm]{\protect\propositionname}
\theoremstyle{plain}
\newtheorem{cor}[thm]{\protect\corollaryname}
\theoremstyle{definition}
\newtheorem{example}[thm]{\protect\examplename}

\usepackage{mathrsfs}
\usepackage{amsfonts}
\usepackage{graphicx}
\usepackage{amsthm}
\usepackage{latexsym}
\usepackage{amssymb}
\usepackage{caption}

\usepackage{geometry}
\usepackage{setspace} 
\makeatother

\providecommand{\corollaryname}{Corollary}
\providecommand{\definitionname}{Definition}
\providecommand{\examplename}{Example}
\providecommand{\factname}{Fact}
\providecommand{\lemmaname}{Lemma}
\providecommand{\propositionname}{Proposition}
\providecommand{\remarkname}{Remark}
\providecommand{\theoremname}{Theorem}

\begin{document}
\title{When Stable and Preferred Semantics Agree: A Path-Based Approach}
\author{Yan Zhang}
\address{Department of Philosophy, Renmin University of China, Beijing,  P.\ R.\ China}
\email{yanzhangsym@gmail.com}
\thanks{Corresponding author: Yan Zhang.}
\author{Kehan Liu}
\address{Department of Philosophy, Renmin University of China, Beijing,  P.\ R.\ China}
\keywords{abstract argumentation, stable semantics, preferred semantics, semantic agreement, well-foundedness}

\begin{abstract}

We investigate structural conditions under which preferred and stable semantics agree in abstract argumentation frameworks. Our approach is based on suitably chosen sets of paths in the attack graph and binary relations induced by them. We establish a general path-based criterion for semantic agreement and instantiate it with several classes of paths, obtaining sufficient conditions expressed in terms of well-foundedness of the corresponding relations.

The resulting conditions strictly generalize two principal structural criteria for preferred and stable agreement in the literature: Dung's limited controversiality condition and the SCC-symmetry condition. In particular, Dung's result is recovered as a special case by taking the relevant path set to be the set of even paths. We further introduce two infinite sequences of path sets, each yielding a family of agreement conditions. For the first sequence, a Collapse Theorem reduces the infinitely many associated well-foundedness conditions to three essentially distinct ones, whereas the second sequence gives rise to genuinely distinct conditions with no analogous collapse. Our results apply to both finite and infinite argumentation frameworks, admit simplified formulations in the finite case, and also yield sufficient conditions for the existence of stable extensions.
\end{abstract}

\maketitle

\section{Introduction}
Dung's theory of abstract argumentation \cite{dung1995acceptability} provides a general framework for representing and evaluating arguments. Among its classical semantics, stable and preferred semantics play a central role. In general, however, the two semantics need not determine the same extensions. This raises a natural structural question: under what conditions do stable and preferred semantics agree?

The agreement problem\footnote{Although frameworks in which preferred and stable semantics agree are often called coherent, we use the broader terminology of semantics agreement, following established usage in the argumentation literature; see, e.g., \cite{baroni2018abstract}.} is significant from both computational and semantic perspectives. For finite abstract argumentation frameworks, deciding whether stable and preferred semantics agree is $\Pi_2^p$-complete \cite{dunne2002coherence}. Hence, unless the polynomial hierarchy collapses, one cannot expect a polynomial-time decision procedure for the problem in general, which makes the search for structural sufficient conditions particularly relevant. Such conditions are also important because they can guarantee the existence of stable extensions. Indeed, every stable extension is preferred, and preferred extensions always exist; consequently, whenever the two semantics agree, a stable extension necessarily exists.

The existence of stable extensions is itself an important problem, with connections extending beyond abstract argumentation. Stable semantics is closely related to two foundational formalisms in nonmonotonic reasoning: the stable-model semantics of logic programming and the extension semantics of default logic \cite{gelfond1988stable,reiter1980logic}. Dung \cite{dung1995acceptability} established fundamental connections between these formalisms and abstract argumentation, and Bondarenko et al.\ \cite{bondarenko1997abstract} developed these connections further in the setting of assumption-based argumentation. Correspondingly, conditions guaranteeing the existence of stable models and default extensions have long been studied in their respective settings \cite{gelfond1988stable,reiter1980logic,papadimitriou1994default}.

Several sufficient conditions for the agreement of preferred and stable semantics are already known. Dung proved that the two semantics agreement for limited controversial argumentation frameworks \cite{dung1995acceptability}. In assumption-based argumentation, related agreement results were obtained for stratified and order-consistent frameworks \cite{bondarenko1997abstract}. For finite abstract argumentation frameworks, symmetry together with the absence of self-attacks guarantees agreement \cite{coste2005symmetric}; this result was later generalized to SCC-symmetric frameworks without self-attacks \cite{baroni2007characterizing}. More recently, \cite{schulz2018responsibility,ijcai2019p892} characterized non-stable preferred labellings in terms of initial strongly connected components consisting of undecided arguments.\footnote{Preferred and stable labellings are in one-to-one correspondence with preferred and stable extensions, respectively; see \cite{caminada2009logical}.} A broader overview can be found in \cite{baroni2018abstract}.

There is also a classical graph-theoretic perspective. An abstract argumentation framework can be viewed as a directed graph, and stable extensions correspond to kernels of the associated digraph. Hence, results on kernel existence therefore yield corresponding results on stable extension existence. Richardson's theorem, for example, implies that every finite framework whose associated digraph contains no odd directed cycle has a stable extension \cite{richardson1953solutions}; see also \cite{goldfeder2025theorem}. Duchet established further sufficient conditions involving symmetric arcs on directed cycles \cite{duchet1980graphes}, and related results have been developed for infinite digraphs \cite{galeana2009some,rojas2010kernels,walicki2019kernels}. 

Our approach is based on paths in the attack graph. We consider abstract classes of path sets and associate with each path set \(\mathcal{P}\) a binary relation \(\prec^{\mathcal{P}}\). This abstraction allows the general argument for semantic agreement to be separated from the particular choice of paths. This leads to the Fundamental Theorem, which underlies the agreement criteria developed throughout the paper.
We then specialize this abstract construction to several natural classes of paths, including even paths and strict even paths, and obtain concrete sufficient conditions for agreement expressed through well-foundedness properties of the corresponding relations. The resulting conditions strictly generalize the two principal structural criteria for preferred and stable agreement discussed above: Dung's limited controversiality condition and the SCC-symmetry condition. In particular, Dung's result is recovered as a special case by taking the relevant path set to be the set of even paths. This approach may therefore be viewed as an abstraction of the path structure underlying Dung's condition, while allowing other choices of path sets that lead to more general agreement criteria.

We develop this construction further through two infinite sequences of path sets
$$
\{\mathcal{P}_n\}_{n\in\mathbb{N}}
\qquad\text{and}\qquad
\{\mathcal{S}_n\}_{n\in\mathbb{N}}.
$$
Each sequence gives rise to a family of well-foundedness conditions sufficient for semantic agreement, but the two families have markedly different structures. For the first sequence, the Collapse Theorem shows that the infinitely many conditions given by the well-foundedness of the relations
$
\{\prec^{\mathcal{P}_n}\}_{n\in\mathbb{N}}
$
reduce to three essentially distinct conditions: the well-foundedness of $\prec^{\mathcal{P}_0}$, $\prec^{\mathcal{P}_1}$, and $\prec^{\mathcal{P}_2}$. By contrast, the well-foundedness conditions arising from
$
\{\prec^{\mathcal{S}_n}\}_{n\in\mathbb{N}}
$
are genuinely distinct and admit no analogous collapse.
In finite argumentation frameworks, these conditions admit simpler structural formulations. We further establish a reduction principle that localizes the global agreement conditions relative to an admissible set \(S\): to ensure that every preferred extension containing \(S\) is stable, it suffices to verify the relevant condition on the arguments that remain undecided relative to \(S\).

The remainder of the paper is organized as follows. Section~\ref{Sect: Abstract Argumentation} introduces the necessary preliminary notions. Section~\ref{Sect: Agreement Theorems} develops the abstract path-based framework, including admissibly backward $2$-extendable path sets and the associated relations $\prec^{\mathcal{P}}$, and establishes the Fundamental Theorem. Section~\ref{Sect: Two sequences of path sets} introduces the two infinite sequences of path sets and derives the corresponding agreement results. Section~\ref{Sect: Agreement Conditions in Finite Argumentations} studies the resulting conditions in finite argumentation frameworks. Section~\ref{Sect: From Global to Local Semantic Agreement} develops and applies a reduction principle that localizes global agreement conditions relative to an admissible set. Finally, Section~\ref{Sect: Conclusion} concludes the paper.

\section{Abstract Argumentation Frameworks}
\label{Sect: Abstract Argumentation}

In this section, we provide a brief overview of Dung's abstract argumentation theory \cite{dung1995acceptability} and introduce the terminology required for the rest of the paper. 
Central to this theory is the concept of an argumentation framework, which consists of a set of arguments and an attack relation defined over them.

\begin{defn}
An\emph{ abstract argumentation framework }(or simply\emph{ argumentation framework}) 
is an ordered pair $\left\langle A,R\right\rangle $,
where $A$ is a set of arguments and $R$ is a binary relation on $A$. 
\end{defn}

Let $AF=\left\langle A,R\right\rangle $ be an argumentation framework.
The relation $R$ is called the \emph{attack relation}. 
The \emph{asymmetric part} of $R$, denoted by $\vec{R}$, is defined by $\vec{R}:=R\setminus R^{-1}=\{(a,b)\in R:(b,a)\notin R\}$.
The \emph{symmetric part} of $R$, denoted by $\overset{\leftrightarrow}{R}$, is defined by $\overset{\leftrightarrow}{R}:=R \cap R^{-1}=\{(a,b)\in R:(b,a)\in R\}$.
Let $a,b \in A$ and $S \subseteq A$.
We usually say \emph{$a$ attacks} $b$ whenever $Rab$. 
Likewise, we say $S$ \emph{attacks} $b$ whenever $Rab$ for some $a\in S$, and $a$ \emph{attacks} $S$ whenever $Rab$ for some $b\in S$. Furthermore, we use $S_{AF}^{+}$ for the set of all the arguments attacked by $S$, i.e., $\{a\in A:\exists b\in S(Rba)\}$,
$S_{AF}^{-}$ for the set of arguments attacking $S$, i.e., $\{a\in A:\exists b\in S(Rab)\}$ and $\mathsf{u}_{AF}(S)$
for the set of \emph{undecided arguments with respect to} $S$, i.e., $A \setminus (S \cup S_{AF}^{+})$. 
For brevity, when $S$ is a singleton $\{a\}$, we write $a_{AF}^{+}$, $a_{AF}^{-}$ and $\mathsf{u}_{AF}(a)$  instead of $\{a\}_{AF}^{+}$, $\{a\}_{AF}^{-}$ and $\mathsf{u}_{AF}(\{a\})$, and may omit the subscript $AF$ when it is clear from the context.

For any $a\in A$, $a$ is \emph{self-defeating} if $Raa$, and 
$AF$ is \emph{free of self-defeating} if all arguments from it are not self-defeating, i.e., $R$ is irreflexive. 
$AF$ is \emph{finite} if $A$ is finite; $AF$ is \emph{finitary} if each argument has only finitely many attackers, i.e., $a^-_{AF}$ is finite for each argument $a \in A$.

Let $\mathscr{R}$ be a binary relation on the argument set $A$. For any subset 
$S \subseteq A$, an argument $a$ is \emph{$\mathscr{R}$-minimal in $S$} if $a \in S$ and there is no $b \in S$ such that $\mathscr{R}ba$.
We also say $a$ is \emph{$\mathscr{R}$-minimal in $AF$} if it is $\mathscr{R}$-minimal in $A$.
An \emph{infinite descending $\mathscr{R}$-chain} is an infinite sequence $\langle x_n \rangle_{n \in \mathbb{N}}$ such that $\mathscr{R}x_{n+1}x_n$ for all $n \in \mathbb{N}$.
The relation $\mathscr{R}$ is \emph{well-founded} if there is no infinite descending $\mathscr{R}$-chains,
or equivalently, if every nonempty subset $S \subseteq A$ contains an $\mathscr{R}$-minimal element.

An argumentation framework is essentially a directed graph in which the arguments are nodes and the attack relation is the edges. 
Thus, we can use standard graph-theoretic notions, such as edges, paths, path length, and cycles, for argumentation frameworks.

\begin{defn}
Let $AF=\left\langle A,R\right\rangle$ be an argumentation framework. 
\begin{itemize}
\item A \emph{path} is a sequence $\langle a_{0},...,a_{n} \rangle$ (or simply $a_{0},...,a_{n}$) of arguments 
such that $n \geqslant 0$ and $Ra_{i}a_{i+1}$ for each $0\leqslant i<n$; 
The integer $n$ is the \emph{length} of the path $a_{0},...,a_{n}$. 
A path from $a$ to $b$ is a path $a_{0},...,a_{n}$ such that $a_{0}=a$ and $a_{n}=b$. 
\item A \emph{cycle} is a path $a_{0},...,a_{n}$ such that $n>0$ and $a_{0}=a_{n}.$ 
\item Let $\boldsymbol{p}_{1} = \langle a_{0},\dots,a_{n} \rangle$ and $\boldsymbol{p}_{2} = \langle b_{0},\dots,b_{m} \rangle$ be two paths such that $Ra_{n}b_{0}$. The \emph{concatenation} of $\boldsymbol{p}_{1}$ and $\boldsymbol{p}_{2}$, denoted by $\boldsymbol{p}_{1} \circ \boldsymbol{p}_{2}$, is defined as the path $\langle a_{0},\dots,a_{n},b_{0},\dots,b_{m} \rangle$.
\end{itemize}
A path (or cycle) of odd length is called an \emph{odd path (or cycle)}, and a path (or cycle) of even length is called an \emph{even path (or cycle)}.

For any $a,b\in A$, we say that $(a,b)$ is a \emph{strict edge} from $a$ to $b$ if $\vec{R}ab$, i.e., if $Rab$ and $\neg Rba$, and that there is a \emph{symmetric edge} between $a$ and $b$ if $\overset{\leftrightarrow}{R}ab$, i.e., if $Rab$ and $Rba$.
A \emph{strict path (or cycle)} in $AF$ is a path (or cycle) consisting entirely of strict edges. A path (or cycle) $a_0,\ldots,a_n$ \emph{traverses a symmetric edge} if there is a symmetric edge between $a_i$ and $a_{i+1}$ for some $0\leqslant i\leqslant n-1$.
\end{defn}

Note that we allow the trivial paths with the length $0$, but all cycles have length greater than $0$.
\begin{defn}
Let $AF=\left\langle A,R\right\rangle $ be an argumentation framework
and $S\subseteq A$. $S$ is \emph{conflict-free }in $AF$ iff there
are no arguments $a$ and $b$ in $S$ such that $Rab$; an argument
$a\in A$ is \emph{acceptable} in $AF$ with respect to $S$ if for
all $b\in A$ such that $Rba$, $b\in S^{+}$. The \emph{characteristic
function} $\mathcal{F}_{AF}:2^{A}\rightarrow2^{A}$ is defined by
taking: 
\[
\mathcal{F}_{AF}(S)=\{a\in A:a\text{ is acceptable in \ensuremath{AF} with respect to }S\}.
\]
\end{defn}

It is easy to see that $\mathcal{F}_{AF}$ is monotonic. In \cite{dung1995acceptability},
Dung proposed four classical semantics for argumentation frameworks:
grounded, complete, stable, and preferred semantics. The ideal semantics
is proposed in \cite{dung2006dialectic}, which is a unique-status approach like the grounded
semantics, but might accept a larger set of arguments. 
\begin{defn}
Let $AF=\left\langle A,R\right\rangle $ be an argumentation framework
and $S\subseteq A$.
\begin{itemize}
    \item $S$ is \emph{admissible }in $AF$\emph{ }if $S$ is conflict-free
in $AF$ and $S\subseteq\mathcal{F}_{AF}(S)$. 
    \item $S$ is a \emph{complete extension} of $AF$\emph{ }if $S$ is admissible
and $\mathcal{F}_{AF}(S)\subseteq S$.
    \item $S$ is a \emph{preferred extension} of $AF$\emph{ }if $S$ is a
maximal admissible set in $AF$ (with respect to the set inclusion).
\item $S$ is a \emph{stable extension} of $AF$ if $S=A-S^{+}$.
    \item $S$ is a \emph{grounded extension} of $AF$ if $S$ is the least
fixed point of $\mathcal{F}_{AF}$.
    \item $S$ is a \emph{ideal extension of} $AF$ if $S$ is a maximal admissible
set in $AF$ such that $S\subseteq E$ for all preferred extensions
$E$.
\end{itemize}
\end{defn}

Obviously, for any conflict-free set $S$, it is a complete extension
iff it is a fixed point of $\mathcal{F}_{AF}$. 
It has been shown in \cite{dung1995acceptability} that every stable extension is a preferred extension. 
However, not all preferred extensions are stable.
Consider the argumentation framework $AF = \left\langle A,R\right\rangle$ depicted in Figure \ref{Fig: no stable extension}, 
where $A = \{a,b\}$ and $R = \{\langle a,a\rangle, \langle a,b\rangle\}$. 
It has no stable extension but has $\textrm{Ø}$ as its sole preferred extension.

\begin{figure}[htbp]
    \centering
    \begin{tikzpicture}[
        argument/.style={
            circle,
            draw,
            thick,
            minimum size=9mm,
            inner sep=0pt
        },
        attack/.style={-{Stealth[length=2.5mm]}, thick},
    ]

    \node[argument] (a) at (0,0) {$a$};
    \node[argument] (b) at (2.5,0) {$b$};

    \draw[attack]
        (a) to[out=120,in=60,looseness=7] (a);

    \draw[attack] (a) -- (b);

    \end{tikzpicture}
    \caption{}
    \label{Fig: no stable extension}
\end{figure}

The following theorem follows directly from Theorem 11 in \cite{dung1995acceptability}.

\begin{thm}
\label{thm: preferred contain admissible set}
Let $AF=\left\langle A,R\right\rangle$
be an argumentation framework and $S$ an admissible set in $AF$.
There exists a preferred extension $E$ of $AF$ such that $S\subseteq E$.
\end{thm}

Finally, we introduce the concept of subframeworks and establish several basic lemmas that will be used in the rest of the paper.

\begin{defn}
Let $AF=\left\langle A,R\right\rangle $ be an argumentation framework
and $S\subseteq A$. 
A \emph{subframework} of $AF$ is an argumentation framework $\left\langle A',R'\right\rangle $ such that $A'\subseteq A$ and $R'= R\cap(S\times S)$. 
The \emph{subframework} of $AF$ \emph{induced} by $S$ (written as $AF{\downarrow}_{S}$) is the subframework of $AF$ with the argument set $S$. 
\end{defn}

\begin{lem}
\label{fact: no attack each other}Let $AF=\left\langle A,R\right\rangle $
be an argumentation framework and $S$ an admissible set in $AF$.
For any $S'\subseteq A$, if $S$ attacks no argument in $S'$,
then $S'$ also doesn't attack any argument in $S$.
\end{lem}
\begin{proof}
Suppose some $a\in S'$ attacks some $b\in S$. Since $S$ is admissible, some $c\in S$ attacks $a$, contradicting that $S$ attacks no argument in $S'$. Hence, $S'$ attacks no argument in $S$.
\end{proof}


\begin{lem}
\label{lem: union of two ad-ext}
Let $AF=\left\langle A,R\right\rangle $ be an argumentation framework, $S$ an admissible set in $AF$ and $S'$ an admissible set in $AF{\downarrow}_{\mathsf{u}(S)}$.
Then $S\cup S'$ is an admissible set in $AF$.
\end{lem}

\begin{proof}
Since $S'\subseteq \mathsf{u}(S)$, $S$ does not attack any argument in $S'$, and thus by Lemma \ref{fact: no attack each other}, $S'$ also does not attack any argument in $S$. 
Hence, $S\cup S'$ is conflict-free. 
Since $S$ is admissible in $AF$, we have $S\subseteq\mathcal{F}_{AF}(S)$,
and hence by monotonicity of $\mathcal{F}_{AF}$, $S\subseteq\mathcal{F}_{AF}(S\cup S')$.
To show that $S'\subseteq\mathcal{F}_{AF}(S\cup S')$, let $a\in S'$ and $b\in A$ such that $Rba$. Because $S\cup S'$ is conflict-free, either $b\in S^{+}$ or $b\in\mathsf{u}(S)$. 
If $b\in\mathsf{u}(S)$, then we obtain by the admissibility of $S'$ in $AF{\downarrow}_{\mathsf{u}(S)}$
that $Rcb$ for some $c\in S'$. Therefore, $S'\subseteq\mathcal{F}_{AF}(S\cup S')$.
\end{proof}

\section{Agreement Theorems}
\label{Sect: Agreement Theorems}
In this section, we develop a unified theoretical framework to establish the agreement between stable and preferred semantics. To this end, we introduce a generic, path-dependent relation parameterized by an arbitrary set of paths. By imposing a natural extendability condition on these path sets, we prove the Fundamental Theorem for semantic agreement. We subsequently demonstrate the utility of this abstraction by instantiating it with specific path sets. This approach yields concrete structural conditions---formulated in terms of well-foundedness---that guarantee every preferred extension is stable.

\begin{defn}
\label{def: generic path relation}
Let $AF=\left\langle A,R\right\rangle $ be an argumentation framework and $\mathcal{P}$ a set of paths in $AF$. We define the relation $\prec_{AF}^{\mathcal{P}}$ on $A$ such that for any $a,b\in A$, $a \prec_{AF}^{\mathcal{P}} b$ if and only if:
\begin{enumerate}
    \item there is a path from $a$ to $b$ in $\mathcal{P}$;
    \item there exists $c\in A$ with $Rac$ such that there is a path from $c$ to $b$ in $\mathcal{P}$.
\end{enumerate}
\end{defn}

By instantiating the generic path set $\mathcal{P}$, we can recover established relations from the literature as well as introduce novel structural conditions:
\begin{example}
Let $AF=\left\langle A,R\right\rangle$ be an argumentation framework, 
let $\mathcal{E}$ (or $\vec{\mathcal{E}}$) denote the set of all paths (or strict paths) of even length in $AF$, and let $\mathcal{O}$ (or $\vec{\mathcal{O}}$) denote the set of all paths (or strict paths) of odd length in $AF$. 

The relation $\prec_{AF}^{\mathcal{E}}$ precisely captures Dung's concept of arguments being \emph{controversial}. Consequently, the well-foundedness of $\prec_{AF}^{\mathcal{E}}$ is equivalent to $AF$ being \emph{limited controversial} \cite{dung1995acceptability}. 

As we will show uniformly in Theorem \ref{thm: the agreement theorem by E and O}, the well-foundedness of either $\prec_{AF}^{{\mathcal{E}}}$, $\prec_{AF}^{\vec{\mathcal{E}}}$, $\prec_{AF}^{{\mathcal{O}}}$ or $\prec_{AF}^{\vec{\mathcal{O}}}$ is sufficient to guarantee the agreement between stable and preferred semantics.
\end{example}

Let $AF=\left\langle A,R\right\rangle $ be an argumentation framework and $\mathcal{P}$ a set of paths in $AF$. To facilitate the discussion of the relation on subframeworks, we adopt the following notational conventions. For any subset $B\subseteq A$, let $\mathcal{P}{\downarrow}_{B}$ denote the restriction of $\mathcal{P}$ to $B$ (i.e., the set of all paths in $\mathcal{P}$ consisting solely of arguments in $B$). We write $\prec_{B}^{\mathcal{P}}$ to denote the relation associated with the subframework $AF{\downarrow}_{B}$ and the restricted path set $\mathcal{P}{\downarrow}_{B}$. Formally:
\[
    \prec_{B}^{\mathcal{P}} \quad \equiv \quad \prec_{AF{\downarrow}_{B}}^{\mathcal{P}{\downarrow}_{B}}.
\]
When the path set $\mathcal{P}$ is fixed by context, we may simply write $\prec_{B}$. In the case where $B=A$, we denote the relation on the full framework interchangeably as $\prec_{A}^{\mathcal{P}}$, $\prec_{AF}^{\mathcal{P}}$, or simply $\prec^{\mathcal{P}}$.

We now establish some basic properties of the general relation $\prec^{\mathcal{P}}$. The following lemma demonstrates that the relation is monotonic with respect to subframework inclusion.

\begin{lem}
\label{lem: not the restriction} \label{lem: monotonicity of path relation w.r.t. argument sets}
Let $AF=\left\langle A,R\right\rangle$ be an argumentation framework and $\mathcal{P}$ a set of paths in $AF$. 
For any $B,C \subseteq A$, if $B \subseteq C$, then ${\prec_{B}} \subseteq {\prec_{C}}$. 
\end{lem}

\begin{proof}
    Let $a,b\in B$. Suppose that $a \prec_{B} b$. Then there is a path $\boldsymbol{p}_{1}$ from $a$ to $b$ in $\mathcal{P}{\downarrow}_B$, and there exists $c \in B$ with $Rac$ such that there is a path $\boldsymbol{p}_{2}$ from $c$ to $b$ in $\mathcal{P}{\downarrow}_B$. Since $B \subseteq C$, $\mathcal{P}{\downarrow}_B \subseteq \mathcal{P}{\downarrow}_C$, thus both $\boldsymbol{p}_{1}$ and $\boldsymbol{p}_{2}$ are paths in $\mathcal{P}{\downarrow}_C$. Hence, $a \prec_{C} b$.
\end{proof}

It is important to note that the relation $\prec_{B}$ is not necessarily the exact restriction of $\prec_{C}$ to the set $B$. While Lemma \ref{lem: monotonicity of path relation w.r.t. argument sets} guarantees that ${\prec_{B}} \subseteq ({\prec_{C}} \cap (B \times B))$, the converse does not generally hold. There may exist arguments $a, b \in B$ such that $a \prec_{C} b$ is true, but $a \prec_{B} b$ is false. This occurs when the paths required to establish the relation in $C$ route through arguments that are excluded from $B$.

\begin{example}
    \label{examine:not-a-restriction}
    Let $AF = \langle A, R \rangle$ as depicted in Figure \ref{fig:triangle-with-tail}, where $A = \{a, b, c, d\}$ and $R = \{\langle a,d\rangle, \langle d,b\rangle, \langle a,c\rangle, \langle c,d\rangle\}$. Let $\mathcal{P}$ be the set of all strict even paths in $AF$. Consider the sets $C = A$ and $B = \{a, b, c\}$. 
    
    In the subframework $AF{\downarrow}_C$, which coincides with $AF$, there is a path $a, d, b$ from $a$ to $b$, and a path $c, d, b$ from $c$ to $b$ with $Rac$. Thus, $a \prec_C b$. 
    
    However, in the restricted subframework $AF{\downarrow}_B$, the argument $d$ is removed. The only remaining attack is $Rac$. Consequently, there is no path from $a$ to $b$ entirely within $B$, meaning $a \prec_B b$ does not hold.
\end{example}

\begin{figure}[htbp]
    \centering
    \begin{tikzpicture}[
        x=1.4cm,
        y=1.2cm,
        argument/.style={
            circle,
            draw,
            thick,
            minimum size=9mm,
            inner sep=0pt
        },
        attack/.style={-{Stealth[length=2.5mm]}, thick},
    ]

        \node[argument] (a) at (0,0)   {$a$};
        \node[argument] (c) at (1.5,1) {$c$};
        \node[argument] (d) at (1.5,-1) {$d$};
        \node[argument] (b) at (3,-1)  {$b$};

        \draw[attack] (a) -- (d);
        \draw[attack] (d) -- (b);
        \draw[attack] (a) -- (c);
        \draw[attack] (c) -- (d);

    \end{tikzpicture}

    \caption{}
    \label{fig:triangle-with-tail}
\end{figure}

The following lemma shows that our general relation is monotonic with respect to path sets inclusion; specifically, it is preserved when restricting the path set to a smaller path set.
\begin{lem}
\label{lem: monotonicity of path relation w.r.t. path sets}
Let $AF=\left\langle A,R\right\rangle$ be an argumentation framework and $\mathcal{P, P'}$ two sets of paths in $AF$. 
If $\mathcal{P} \subseteq \mathcal{P'}$, then ${\prec^{\mathcal{P}}} \subseteq {\prec^{\mathcal{P'}}}$. 
\end{lem}
\begin{proof}
    Let $a,b \in A$ and suppose that $a \prec^{\mathcal{P}} b$. By the definition of $\prec^{\mathcal{P}}$, there is a path $\boldsymbol{p}_{1}$ from $a$ to $b$ in $\mathcal{P}$, and there exists an argument $c \in A$ with $Rac$ such that there is a path $\boldsymbol{p}_{2}$ from $c$ to $b$ in $\mathcal{P}$. Since $\mathcal{P} \subseteq \mathcal{P'}$, both paths $\boldsymbol{p}_{1}$ and $\boldsymbol{p}_{2}$ are also contained in $\mathcal{P'}$. Consequently, there is a path from $a$ to $b$ in $\mathcal{P'}$, and a path from $c$ to $b$ in $\mathcal{P'}$ with $Rac$. This satisfies the definition of $a \prec^{\mathcal{P'}} b$. Therefore, ${\prec^{\mathcal{P}}} \subseteq {\prec^{\mathcal{P'}}}$.
\end{proof}

In order to guarantee the existence of nonempty admissible sets within our general framework, we require the path set $\mathcal{P}$ to satisfy a specific extendable property.
\begin{defn}
\label{def: path set properties}
Let $AF=\left\langle A,R\right\rangle$ be an argumentation framework, $B \subseteq A$, and $\mathcal{P}$ a set of paths in $AF$.
\begin{itemize}
    \item We say $\mathcal{P}$ is \emph{closed under strict $2$-prepending} if, for every path $\boldsymbol{p} \in \mathcal{P}$ originating at an argument $a \in A$, and any pair of arguments $b, c \in A$ such that $\vec{R}cb$ and $\vec{R}ba$, the concatenated path $\langle c, b \rangle \circ \boldsymbol{p}$ also belongs to $\mathcal{P}$.
    \item We say $\mathcal{P}$ is \emph{strictly backward $2$-extendable with respect to} $B$ if, for every path $\boldsymbol{p} \in \mathcal{P}{\downarrow}_B$ originating at some $a \in B$ and any strict attacker $b \in B$ (i.e., $\vec{R}ba$) of $a$, the existence of some $c \in B$ satisfying $\vec{R}cb$ guarantees the existence of a (possibly distinct) $c' \in B$ such that $\langle c', b \rangle \circ \boldsymbol{p}$ is also contained in $\mathcal{P}{\downarrow}_B$.
    \item We say $\mathcal{P}$ is \emph{admissibly backward $2$-extendable} if, for every admissible set $S \subseteq A$, $\mathcal{P}$ is strictly backward $2$-extendable with respect to the corresponding set of undecided arguments, $\mathsf{u}(S)= A \setminus (S \cup S^+)$.

\end{itemize}
\end{defn}

Note that closure under strict 2-prepending constitutes a global, structural property that trivially implies strict backward $2$-extendability with respect to any arbitrary subset $B \subseteq A$. Consequently, any path set closed under strict 2-prepending is inherently admissibly backward $2$-extendable; the converse, however, fails to hold in general. As we will subsequently demonstrate, our fundamental agreement theorem relies on this weaker requirement of admissible backward extendability. Furthermore, one may readily verify that the parity-constrained path sets $\mathcal{E}$, $\vec{\mathcal{E}}$, $\mathcal{O}$, and $\vec{\mathcal{O}}$ are all automatically closed under strict 2-prepending, thereby satisfying the required extendability condition across all undecided argument sets.

The following lemma establishes a sufficient condition for the existence of nonempty admissible sets based on the minimality with respect to the relation $\prec^{\mathcal{P}}$.

\begin{lem}
\label{lem: existence of nonempty admissible set}
    Let $AF=\left\langle A,R\right\rangle$ be an argumentation framework that is free of self-defeating and $\mathcal{P}$ a set of paths in it that is strictly backward 2-extendable with respect to $A$.
    Suppose that
    \begin{enumerate}
        \item either $A$ contains an $\vec{R}$-minimal element,
        \item or there exists a path in $\mathcal{P}$ ending at a $\prec^{\mathcal{P}}$-minimal element in $A$.
    \end{enumerate}
    Then $AF$ admits a nonempty admissible set.
\end{lem}

\begin{proof}
    


    First, consider the case where $A$ contains an $\vec{R}$-minimal element, say $x$. Since $AF$ is free of self-defeating, the set $\{x\}$ is conflict-free. Furthermore, for any $y \in A$ such that $Ryx$, the $\vec{R}$-minimality of $x$ implies that $\neg \vec{R}yx$, and thus we must have $Rxy$. Consequently, $\{x\}$ defends itself against all attacks and is a nonempty admissible set.
    
    Now, assume that $AF$ contains no $\vec{R}$-minimal element. Then there exists path  in $\mathcal{P}$ ending at some $\prec^{\mathcal{P}}$-minimal element in $A$, say $a$. Define the set
    \begin{equation*}
        S = \{b\in A \mid \text{there is a path from } b \text{ to } a \text{ in } \mathcal{P}\}.
    \end{equation*}
    Since some path in $\mathcal{P}$ ends at $a$, $S$ is nonempty. 

    We claim that $S$ is conflict-free. Suppose, for the sake of contradiction, that there exist $b, c \in S$ such that $Rbc$. By the definition of $S$, there is a path from $c$ to $a$ in $\mathcal{P}$. The existence of this path, combined with the attack $Rbc$ and the fact that $b \in S$ (so there is a path from $b$ to $a$ in $\mathcal{P}$), implies that $b \prec^{\mathcal{P}} a$. This contradicts the minimality of $a$. Thus, $S$ is conflict-free.
    
    It remains to show that $S\subseteq\mathcal{F}_{AF}(S)$. Let $b \in S$ and suppose $c \in A$ attacks $b$ (i.e., $Rcb$). We must show that $S$ attacks $c$. 
    If $Rbc$, then $b$ itself attacks $c$, and since $b \in S$, the condition is satisfied.
    Suppose $\neg Rbc$. Then we have a strict attack $\vec{R}cb$. Since $A$ has no $\vec{R}$-minimal elements, there exists some $d \in A$ such that $\vec{R}dc$.
    We now have $d, c \in A$ such that $\vec{R}dc$ and $\vec{R}cb$. Since $b \in S$, there is a path $\boldsymbol{p}$ from $b$ to $a$ in $\mathcal{P}$. Because $\mathcal{P}$ is strictly backward 2-extendable with respect to $A$, there is a $d' \in A$ such that the path formed by concatenating $\langle d', c \rangle $ and $ \boldsymbol{p}$ is also in $\mathcal{P}$. Therefore, $d' \in S$. Since the existence of this path implies $Rd'c$, the set $S$ attacks $c$.
    
    From the above, we conclude that $S$ is a nonempty admissible set in $AF$.
\end{proof}

We now present our fundamental agreement theorem, formulated in terms of the general path relation $\prec$. While abstract, this result serves as a unified foundation for deriving concrete agreement conditions based on specific path properties.

\begin{thm}[Fundamental Theorem]
\label{thm: the fundamental agreement theorem}
    Let $AF=\left\langle A,R\right\rangle$ be an argumentation framework that is free of self-defeating, and let $\mathcal{P}$ be an admissibly backward $2$-extendable set of paths in $AF$.
    Suppose that for every admissible set $S\subseteq A$ with $\mathsf{u}(S)\neq\emptyset$, 
    \begin{enumerate}
        \item  either $\mathsf{u}(S)$ contains an $\vec{R}$-minimal element, 
        \item or there is a path in $\mathcal{P}{\downarrow}_{\mathsf{u}(S)}$ ending at a $\prec^{\mathcal{P}}_{\mathsf{u}(S)}$-minimal element of $\mathsf{u}(S)$.
    \end{enumerate}
    Then every preferred extension of $AF$ is stable.
\end{thm}

\begin{proof}
    Suppose, for the sake of contradiction, that there exists a preferred extension $E$ of $AF$ which is not stable. The corresponding set of undecided arguments, $\mathsf{u}(E) = A \setminus (E \cup E^+)$, is therefore nonempty. Since $\mathcal{P}$ is admissibly backward $2$-extendable, $\mathcal{P}$ is strictly backward 2-extendable with respect to $\mathsf{u}(E)$, and thus $\mathcal{P}{\downarrow}_{\mathsf{u}(E)}$ is strictly backward 2-extendable with respect to $\mathsf{u}(E)$.
    By the hypothesis of the theorem, $\mathsf{u}(E)$ either contains an $\vec{R}$-minimal element, 
    or there exists a path in $\mathcal{P}{\downarrow}_{\mathsf{u}(E)}$ ending at a $\prec_{\mathsf{u}(E)}^{\mathcal{P}}$-minimal element in $\mathsf{u}(E)$. 
    Thus, the subframework $AF{\downarrow}_{\mathsf{u}(E)}$ satisfies the conditions of Lemma \ref{lem: existence of nonempty admissible set}. 
    Consequently, there exists a nonempty set $S \subseteq \mathsf{u}(E)$ that is admissible in $AF{\downarrow}_{\mathsf{u}(E)}$.
    It follows from Lemma \ref{lem: union of two ad-ext} that the union $E \cup S$ is an admissible set in $AF$. Since $S$ is nonempty and disjoint from $E$, this strictly larger set contradicts the maximality of $E$ as a preferred extension.
    Thus, we conclude that every preferred extension of $AF$ must be stable.
\end{proof}

\begin{rem}
    Let  ${\prec^{\mathcal{P}}}{\downarrow}_{\mathsf{u}(S)} = {\prec^{\mathcal{P}}} \cap (\mathsf{u}(S) \times \mathsf{u}(S))$. By Lemma \ref{lem: monotonicity of path relation w.r.t. argument sets}, ${\prec^{\mathcal{P}}_{\mathsf{u}(S)}} \subseteq {\prec^{\mathcal{P}}}{\downarrow}_{\mathsf{u}(S)}$, 
    and hence condition (2) in Theorem \ref{thm: the fundamental agreement theorem} can be replaced by
    \begin{itemize}
        \item there is a path in $\mathcal{P}{\downarrow}_{\mathsf{u}(S)}$ ending at a ${\prec^{\mathcal{P}}}{\downarrow}_{\mathsf{u}(S)}$-minimal element of $\mathsf{u}(S)$.
    \end{itemize}
    However, this condition is generally stronger than condition (2) in Theorem \ref{thm: the fundamental agreement theorem}, 
    since ${\prec^{\mathcal{P}}_{\mathsf{u}(S)}}$ may be a proper subset of ${\prec^{\mathcal{P}}}{\downarrow}_{\mathsf{u}(S)}$, as illustrated in Example \ref{examine:not-a-restriction}. We have nevertheless adopted the weaker condition to obtain a stronger theorem, even though our subsequent development requires only the weaker version obtained by replacing ${\prec^{\mathcal{P}}_{\mathsf{u}(S)}}$ with ${\prec^{\mathcal{P}}}{\downarrow}_{\mathsf{u}(S)}$ in condition (2).
\end{rem}

By applying Theorem \ref{thm: the fundamental agreement theorem} to the specific path sets introduced previously, we obtain concrete structural conditions for semantics agreement. The following theorem demonstrates that the well-foundedness of the relation induced by either even paths, odd paths, strict even paths, or strict odd paths is sufficient to guarantee that all preferred extensions are stable. Since the well-foundedness of $\prec_{AF}^{\mathcal{E}}$ is equivalent to $AF$ being \emph{limited controversial}, the following theorem recovers Theorem 33 of \cite{dung1995acceptability}.

\begin{thm}
\label{thm: the agreement theorem by E and O}
    Let $AF=\left\langle A,R\right\rangle$ be an argumentation framework that is free of self-defeating and $\mathcal{P} \in \{\mathcal{E}, \mathcal{O}, \vec{\mathcal{E}}, \vec{\mathcal{O}}\}$.
    Suppose that $\prec^{\mathcal{P}}$ is well-founded. Then all preferred extensions of $AF$ are stable.
\end{thm}
\begin{proof}
We will show that the conditions of Theorem \ref{thm: the fundamental agreement theorem} are satisfied.
    First, the path sets $\mathcal{E}, \mathcal{O}, \vec{\mathcal{E}}$, and $\vec{\mathcal{O}}$ are all closed under strict 2-prepending, which implies they are admissibly backward 2-extendable.
    
    Let $S \subseteq A$ be an admissible set such that $\mathsf{u}(S)$ is nonempty. Because $\prec^{\mathcal{P}}$ is well-founded, $\prec^{\mathcal{P}}_{\mathsf{u}(S)}$ is well-founded (by Lemma \ref{lem: monotonicity of path relation w.r.t. argument sets}). Thus, $\mathsf{u}(S)$ contains a $\prec^{\mathcal{P}}_{\mathsf{u}(S)}$-minimal element, say $x$.
    
    We consider the possible instantiations of $\mathcal{P}$:
    \begin{itemize}
        \item Case 1: $\mathcal{P} \in \{\mathcal{E}, \vec{\mathcal{E}}\}$. The trivial path of length 0 from $x$ to $x$ is an even path consisting entirely of elements in $\mathsf{u}(S)$, and thus belongs to $\mathcal{P}{\downarrow}_{\mathsf{u}(S)}$. Therefore, there exists a path in $\mathcal{P}{\downarrow}_{\mathsf{u}(S)}$ ending at $x$, satisfying condition (2) of Theorem \ref{thm: the fundamental agreement theorem}.
        \item Case 2: $\mathcal{P} \in \{\mathcal{O}, \vec{\mathcal{O}}\}$. If there exists a path in $\mathcal{P}{\downarrow}_{\mathsf{u}(S)}$ ending at $x$, then condition (2) is satisfied. If there is no path in $\mathcal{P}{\downarrow}_{\mathsf{u}(S)}$ ending at $x$, then, since $\mathcal{P}{\downarrow}_{\mathsf{u}(S)}$ contains all strict paths of length 1 inside $\mathsf{u}(S)$, there is no $y \in \mathsf{u}(S)$ such that $\vec{R}yx$. This means $x$ is an $\vec{R}$-minimal element in $AF{\downarrow}_{\mathsf{u}(S)}$, satisfying condition (1) of Theorem \ref{thm: the fundamental agreement theorem}.
    \end{itemize}
    
    In all cases, the hypothesis of Theorem \ref{thm: the fundamental agreement theorem} holds. Consequently, all preferred extensions of $AF$ are stable.
\end{proof}

The following corollaries present two direct consequences of the agreement theorem, the second of which recovers Theorem 23 of \cite{galeana2009some}.

\begin{cor}
    Let $AF=\left\langle A,R\right\rangle$ be an argumentation framework that is free of self-defeating and $\mathcal{P} \in \{\mathcal{E}, \mathcal{O}, \vec{\mathcal{E}}, \vec{\mathcal{O}}\}$.
    Suppose that $\prec^{\mathcal{P}}$ is well-founded. Then there is a stable extension of $AF$.
\end{cor}
\begin{proof}
Obviously, $\textrm{Ø}$ is an admissible extension of $AF$, and hence by Theorem \ref{thm: preferred contain admissible set}, there is a preferred extension $E$ of $AF$, which yields by Theorem \ref{thm: the agreement theorem by E and O} that $E$ is also a stable extension.
\end{proof}

In what follows, we shall not state and prove separately the existence of a stable extension in analogous situations. Instead, we incorporate this assertion directly into the corresponding agreement result, as it follows immediately from that result together with the existence of a preferred extension.

An argumentation framework $AF=\langle A,R\rangle$ is \emph{bipartite} if there exists a partition $(X,Y)$ of $A$ such that every attack $\langle a,b\rangle \in R$ consists of one argument from $X$ and one from $Y$.
\begin{cor}
    Let $AF$ be a bipartite argumentation framework. Then all preferred extensions of $AF$ are stable, and hence there is a stable extension of $AF$.
\end{cor}
\begin{proof}
    Since $AF$ is bipartite, it is free of self-defeating.
    Assume that $a \prec^{\mathcal{E}} b$ for some $a,b$ in $AF$.
    Then there is a path from $a$ to $b$ of both even length and odd length.
    But this is impossible in a bipartite argumentation framework.
    Therefore, ${\prec^{\mathcal{E}}}$ is empty, which yields by Theorem \ref{thm: the agreement theorem by E and O} that all preferred extensions of $AF$ are stable.
\end{proof}

\section{Two sequences of path sets}
\label{Sect: Two sequences of path sets}
Having established agreement theorems for individual path sets, we now develop this approach systematically by introducing two infinite sequences of path sets.\footnote{Strictly speaking, the two sequences are defined relative to a fixed argumentation framework and therefore depend on the framework under consideration. This dependence is made explicit in the formal definitions but is elsewhere suppressed for notational convenience, except where required by context.} Each path set $\mathcal{P}$ occurring in
either sequence gives rise to a sufficient condition for the agreement of preferred and stable semantics, namely, the well-foundedness of $\prec^{\mathcal{P}}$. Thus, the two sequences provide two infinite families of sufficient conditions for this agreement. We begin with the first sequence.

\begin{defn}
Let $AF$ be an argumentation framework and let $n \in \mathbb{N}$.
We denote by $\mathcal{P}_n(AF)$ the smallest set of strict paths in
$AF$ such that:
\begin{enumerate}
    \item it contains every strict path in $AF$ of
          length $n$; and
    \item it is closed under strict $2$-prepending.
\end{enumerate}
When $AF$ is clear from the context, we write $\mathcal{P}_n$ in place
of $\mathcal{P}_n(AF)$.
\end{defn}

By definition, because prepending two strict edges increases a path's length by exactly 2 and preserves its parity, $\mathcal{P}_n$ consists precisely of all strict paths of length $l$ such that $l \ge n$ and $l \equiv n \pmod 2$. Consequently, for all $i \ge j$ of the same parity, we have $\mathcal{P}_{i} \subseteq \mathcal{P}_{j}$. By Lemma \ref{lem: monotonicity of path relation w.r.t. path sets}, this inclusion yields ${\prec^{\mathcal{P}_i}} \subseteq {\prec^{\mathcal{P}_j}}$.

\begin{thm}
\label{thm: the agreement theorem by the first sequence}
    Let $AF=\left\langle A,R\right\rangle$ be an argumentation framework that is free of self-defeating and $n \in \mathbb{N}$.
    Suppose that $\prec^{\mathcal{P}_n}$ is well-founded. Then all preferred extensions of $AF$ are stable, and hence there is a stable extension of $AF$.
\end{thm}

\begin{proof}
    By definition, $\mathcal{P}_n$ is closed under strict 2-prepending, hence it is admissibly backward 2-extendable.

    We will verify the conditions of Theorem \ref{thm: the fundamental agreement theorem}. Let $S \subseteq A$ be an admissible set such that $\mathsf{u}(S)$ is nonempty. Because $\prec^{\mathcal{P}_n}$ is well-founded on $A$, $\prec^{\mathcal{P}_n}_{\mathsf{u}(S)}$ is also well-founded (by Lemma \ref{lem: monotonicity of path relation w.r.t. argument sets}), guaranteeing there is a $\prec^{\mathcal{P}_n}_{\mathsf{u}(S)}$-minimal element $x$ in $\mathsf{u}(S)$. 

    Assume that $\mathsf{u}(S)$ contains no $\vec{R}$-minimal element. Then every argument $a$ in $\mathsf{u}(S)$ admits at least one strict predecessor $b$ (i.e., $\vec{R}ba$) in $\mathsf{u}(S)$. This enables us to construct an infinite descending $\vec{R}$-chain originating from $x$ and confined entirely to the subframework $AF{\downarrow}_{\mathsf{u}(S)}$. By traversing this chain, we can trivially obtain a strict path of length $n$ located entirely within $\mathsf{u}(S)$ that ends at $x$, satisfying condition (2) of Theorem \ref{thm: the fundamental agreement theorem}. 

    Therefore, the hypothesis of Theorem \ref{thm: the fundamental agreement theorem} is met, allowing us to conclude that all preferred extensions of $AF$ are stable.
\end{proof}

Consequently, the well-foundedness of any $\prec^{\mathcal{P}_n}$ constitutes a sufficient condition for semantics agreement. The following proposition shows that the well-foundedness of $\vec{R}$ is a sufficient condition for the well-foundedness of every $\prec^{\mathcal{P}_n}$ with $n > 0$. Hence, it also constitutes a sufficient condition for semantics agreement.

\begin{prop}
\label{prop: sufficency of vecR}
    Let $AF=\left\langle A,R\right\rangle$ be an argumentation framework. If $\vec{R}$ is well-founded, then $\prec^{\mathcal{P}_n}$ is well-founded for every $n > 0$.
\end{prop}
\begin{proof}
    Let $n > 0$. We proceed by contraposition. Suppose that the relation $\prec^{\mathcal{P}_n}$ is not well-founded. By definition, there exists an infinite descending $\prec^{\mathcal{P}_n}$-sequence. This sequence naturally induces an infinite descending $\vec{R}$-chain, thereby demonstrating that $\vec{R}$ also fails to be well-founded. 
\end{proof}

The following corollaries present several direct consequences of the agreement theorem for possibly infinite argumentation frameworks.
\begin{cor}
    \label{thm :R vec is well-founded}
    Let $AF=\left\langle A,R\right\rangle$ be an argumentation framework that is free of self-defeating.
    If $\vec{R}$ is well-founded, then all preferred extensions of $AF$ are stable, and hence there is a stable extension of $AF$.
\end{cor}
\begin{proof}
    Apply Theorem \ref{thm: the agreement theorem by the first sequence} and Proposition \ref{prop: sufficency of vecR}.
\end{proof}

The preceding corollary immediately yields the following two further corollaries, which recover Proposition~5 of \cite{coste2005symmetric} and Theorem~2.2 of \cite{rojas2010kernels}, respectively.

\begin{cor}
    Let $AF = \langle A, R \rangle$ be an argumentation framework that is free of self-defeating. 
    If $R$ is symmetrical, then all preferred extensions of $AF$ are stable, and hence there is a stable extension of $AF$.
\end{cor}
\begin{proof}
Since $R$ is symmetrical, $\vec{R} = \textrm{Ø}$, and the result follows directly from Corollary \ref{thm :R vec is well-founded}. 
\end{proof}

\begin{cor}
Let $AF = \langle A, R \rangle$ be an argumentation framework that is free of self-defeating. 
Suppose that every cycle and every infinite descending $\mathscr{R}$-chain of pairwise distinct arguments in it traverses a symmetric edge.
Then all preferred extensions of $AF$ are stable, and hence there is a stable extension of $AF$.
\end{cor}
\begin{proof}
The supposition is an equivalent formulation of the well-foundedness of $\vec{R}$, and hence the result follows directly from Corollary \ref{thm :R vec is well-founded}.
\end{proof}

Although the parameterization by $n$ provides an infinite sequence of sufficient conditions, it raises an obvious question: are these conditions truly distinct? In what follows, we demonstrate that they dramatically collapse into three conditions.

\begin{lem}
\label{lem: Properties of relation based on E2 and O}
    Let $AF=\left\langle A,R\right\rangle $
    be an argumentation framework, and $a,b,c,d \in A$. We have that:
    \begin{enumerate}
        \item \label{eum: lem: Properties of relation based on E2 and O (1)}if $ a \prec^{\mathcal{P}_1} b \prec^{\mathcal{P}_1} c$, then $a \prec^{\mathcal{P}_2} c$;
        \item \label{eum: lem: Properties of relation based on E2 and O (2)}if $ a \prec^{\mathcal{P}_1} b \prec^{\mathcal{P}_1} c \prec^{\mathcal{P}_1} d$, then $a \prec^{\mathcal{P}_1} d$.
    \end{enumerate}
\end{lem}

\begin{proof}
    (\ref{eum: lem: Properties of relation based on E2 and O (1)}) 
    Suppose $a \prec^{\mathcal{P}_1} b$ and $b \prec^{\mathcal{P}_1} c$. By definition, there exist paths $\boldsymbol{p}_{1}, \boldsymbol{p}_{1}' \in \mathcal{P}_1$ terminating at $b$, where $\boldsymbol{p}_{1}$ originates from $a$, and $\boldsymbol{p}_{1}'$ originates from some $b'$ such that $Rab'$. Furthermore, the relation $b \prec^{\mathcal{P}_1} c$ yields a path $\boldsymbol{p}_{2} \in \mathcal{P}_1$ from $b$ to $c$. The concatenation of $\boldsymbol{p}_{1}$ and $\boldsymbol{p}_{2}$ forms a path in $\mathcal{P}_2$ from $a$ to $c$. By concatenating $\boldsymbol{p}_{1}'$ and $\boldsymbol{p}_{2}$, we obtain a path in $\mathcal{P}_2$ from $b'$ to $c$. Since $Rab'$, the existence of these composite paths immediately establishes $a \prec^{\mathcal{P}_2} c$.

    (\ref{eum: lem: Properties of relation based on E2 and O (2)})
    The argument closely mirrors that of (\ref{eum: lem: Properties of relation based on E2 and O (1)}). Unpacking the chained relations yields respective paths in $\mathcal{P}_1$, and concatenating three successive odd paths in $\mathcal{P}_1$ yields another odd path in $\mathcal{P}_1$. 
\end{proof}

\begin{thm}[Collapse Theorem]
\label{thm: collapse theorem}
    Let $AF=\left\langle A,R\right\rangle$ be an argumentation framework. We have:
    \begin{enumerate}
        \item \label{thm: collapse theorem: eum 1} 
        For any odd $n \ge 1$, $\prec^{\mathcal{P}_n}$ is well-founded iff $\prec^{\mathcal{P}_1}$ is well-founded.
        \item \label{thm: collapse theorem: eum 2}
        For any even $n \ge 2$, $\prec^{\mathcal{P}_n}$ is well-founded iff $\prec^{\mathcal{P}_2}$ is well-founded.
        \item \label{thm: collapse theorem: eum 3}
        The well-foundedness of $\prec^{\mathcal{P}_0}$  implies the well-foundedness of $\prec^{\mathcal{P}_2}$.
        \item \label{thm: collapse theorem: eum 4}\label{item: collapse theorem 4} 
        The well-foundedness of $\prec^{\mathcal{P}_2}$  implies the well-foundedness of $\prec^{\mathcal{P}_1}$.
    \end{enumerate}
\end{thm}

\begin{proof}
    First, because $\mathcal{P}_i \subseteq \mathcal{P}_j$ for all $i \ge j$ with the same parity, it follows directly from Lemma \ref{lem: monotonicity of path relation w.r.t. path sets} that ${\prec^{\mathcal{P}_i}} \subseteq {\prec}^{\mathcal{P}_{j}}$. Consequently, the well-foundedness propagates forward with the same parity: the well-foundedness of $\prec^{\mathcal{P}_{n}}$ implies the well-foundedness $\prec^{\mathcal{P}_{n+2k}}$ for all $n, k \ge 0$. This gives the right to the left direction in (\ref{thm: collapse theorem: eum 1}), (\ref{thm: collapse theorem: eum 2}) and the implication of (\ref{thm: collapse theorem: eum 3}).

    (\ref{thm: collapse theorem: eum 1}) Suppose that $\prec^{\mathcal{P}_1}$ is not well-founded. This yields an infinite $\prec^{\mathcal{P}_1}$-descending chain 
    \[
        x_0 \succ^{\mathcal{P}_1} x_1 \succ^{\mathcal{P}_1} x_2 \succ^{\mathcal{P}_1} \cdots.
    \]
    For any odd integer $n$, composing $n$ successive steps along this chain establishes that $x_i \succ^{\mathcal{P}_n} x_{i+n}$. To see this, we must construct the required odd and even paths from $x_{i+n}$ to $x_{i}$. A strict odd path from $x_{i+n}$ to $x_{i}$ is obtained by concatenating $n$ successive strict paths in $\mathcal{P}_1$, each traversing from $x_{j+1}$ to $x_{j}$ for $i \le j < i+n$. An even path from $x_{i+n}$ to $x_{i}$ is formed by adjoining the initial even path from $x_{i+1}$ to $x_{i}$ with the subsequent $n-1$ strict paths in $\mathcal{P}_1$, each traversing from $x_{j+1}$ to $x_{j}$ for $i+1 \le j < i+n$. 

    Consequently, $x_i \succ^{\mathcal{P}_n} x_{i+n}$ holds for all $i \ge 0$. This yields the infinite $\prec^{\mathcal{P}_n}$-descending chain
    \[
        x_0 \succ^{\mathcal{P}_n} x_n \succ^{\mathcal{P}_n} x_{2n} \succ^{\mathcal{P}_n} \cdots,
    \]
    which demonstrates that $\prec^{\mathcal{P}_n}$ is not well-founded.

    (\ref{thm: collapse theorem: eum 2}) Suppose that $\prec^{\mathcal{P}_2}$ is not well-founded. Then there is an infinite $\prec^{\mathcal{P}_2}$-descending chain:
    \[
        x_0 \succ^{\mathcal{P}_2} x_1 \succ^{\mathcal{P}_2} x_2 \succ^{\mathcal{P}_2} \cdots.
    \]
    By a similar argument as (\ref{thm: collapse theorem: eum 1}), i.e., composing $n/2$ successive steps along this chain, we can establish $x_i \succ^{\mathcal{P}_n} x_{i+n/2}$ for all $i$. It follows immediately that 
    \[
        x_0 \succ^{\mathcal{P}_n} x_{n/2} \succ^{\mathcal{P}_n} x_{n} \succ^{\mathcal{P}_n} \cdots
    \]
    forms an infinite $\prec^{\mathcal{P}_n}$-descending chain, confirming that $\prec^{\mathcal{P}_n}$ is not well-founded.

    (\ref{thm: collapse theorem: eum 3}) The implication follows from the fact that the well-foundedness propagates forward with the same parity. 

    (\ref{thm: collapse theorem: eum 4}) Assume that $\prec^{\mathcal{P}_1}$ isn't well-founded. 
    Then there exists an infinite $\prec^{\mathcal{P}_1}$-descending chain:
    \[
        x_0 \succ^{\mathcal{P}_1} x_1 \succ^{\mathcal{P}_1} x_2 \succ^{\mathcal{P}_1} \cdots.
    \]
    By Lemma \ref{lem: Properties of relation based on E2 and O},  $x_0 \succ^{\mathcal{P}_2} x_2\succ^{\mathcal{P}_2}  \cdots x_n\succ^{\mathcal{P}_2} x_{n+2}  \succ^{\mathcal{P}_2} \cdots$. 
    Hence, $\prec^{\mathcal{P}_2}$ isn't well-founded.
\end{proof}

\begin{rem}  
The Collapse Theorem shows that the infinitely many well-foundedness assertions associated with the sequence
$\{\prec^{\mathcal{P}_n}\}_{n\in\mathbb{N}}$
reduce to at most three conditions: the well-foundedness of
$\prec^{\mathcal{P}_0}$, $\prec^{\mathcal{P}_1}$, and
$\prec^{\mathcal{P}_2}$. 
In the next section, we obtain equivalent characterizations of these well-foundedness conditions for finite argumentation frameworks. 
These characterizations provide simple criteria for determining whether a given condition holds, thereby making counterexamples easy to verify. 
In particular, Examples~\ref{example: p2 but not p0}
and~\ref{example: p1 but not p2} show that the implications
in~(\ref{thm: collapse theorem: eum 3}) and~(\ref{thm: collapse theorem: eum 4}), respectively, cannot be reversed. 
Consequently, the three conditions identified by the Collapse
Theorem are genuinely distinct, and hence the collapse to three
conditions is exact.
\end{rem}

The remainder of this section introduces the second sequence of path sets $\left\{\mathcal{S}_n\right\}_{n\in\mathbb{N}}$. As with the first sequence, the well-foundedness of $\prec^{\mathcal{S}_n}$, for any $n\in\mathbb{N}$, is sufficient to ensure that preferred and stable semantics agree. 
In contrast to the well-foundedness conditions associated with the first sequence $\left\{\mathcal{P}_n\right\}_{n\in\mathbb{N}}$, however, those associated with $\left\{\mathcal{S}_n\right\}_{n\in\mathbb{N}}$ are genuinely distinct and do not collapse.

Before defining the sequence, we need some preliminary notions. First, recall that we use $x^-$ for the set of all attackers of $x$, that is, $\{y : Ryx\}$.
\begin{defn}
    Let $AF = \langle A, R \rangle$ be an argumentation framework. We define the \emph{attacker-inclusion preorder} $\sqsubseteq_{AF}$ on $A$ by
    \[
        x \sqsubseteq_{AF} y \iff x^- \subseteq y^-.
    \]
   As usual, we define $x \sqsubset_{AF} y$ iff  $x \sqsubseteq_{AF} y$ but not $y \sqsubseteq_{AF} x$, which is equivalent to $x^- \subset y^-$. When the framework $AF$ is clear from context, we omit the subscript $AF$ in $\sqsubseteq_{AF}, \sqsubset_{AF}$.
    
    A path $x_0, x_1, \ldots, x_n$ in $AF$ is said to be \emph{$\sqsubset$-minimal} if, for every $0 \leq i < n$, the argument $x_{i}$ is a $\sqsubset$-minimal element in $x_{i+1}^-$.

    A set of paths $\mathcal{P}$ in $AF$ is \emph{closed under $\sqsubset$-minimal $2$-prepending} if the following holds: for any path $\boldsymbol{p} \in \mathcal{P}$ originating at an argument $a \in A$, and all arguments $b, c \in A$ such that $\vec{R}cb$ and $\vec{R}ba$, if $c$ is $\sqsubset$-minimal in $b^-$, then the concatenated path $\langle c, b \rangle \circ \boldsymbol{p}$ also belongs to $\mathcal{P}$.
\end{defn}

\begin{defn}
Let $AF$ be an argumentation framework and let $n \in \mathbb{N}$.
We denote by $\mathcal{S}_n(AF)$ the smallest set of strict paths in
$AF$ such that:
\begin{enumerate}
    \item it contains every $\sqsubset$-minimal strict path in $AF$ of
          length $n$; and
    \item it is closed under $\sqsubset$-minimal $2$-prepending.
\end{enumerate}
When $AF$ is clear from the context, we write $\mathcal{S}_n$ in place
of $\mathcal{S}_n(AF)$.
\end{defn}

As with $\mathcal{P}_i$, we have that $\mathcal{S}_i \subseteq \mathcal{S}_j$ for all $i \ge j$ with the same parity, it then follows directly from Lemma \ref{lem: monotonicity of path relation w.r.t. path sets} that ${\prec^{\mathcal{S}_i}} \subseteq {\prec}^{\mathcal{S}_{j}}$. Consequently, the well-foundedness propagates forward with the same parity: the well-foundedness of $\prec^{\mathcal{S}_{n}}$ implies that of $\prec^{\mathcal{S}_{n+2}}$ for all $n \ge 0$.
In contrast to the first sequence, whose well-foundedness conditions collapse to three distinct cases, those arising from
$\left\{\mathcal{S}_n\right\}_{n\in\mathbb{N}}$ are all distinct. This will be demonstrated in Examples \ref{example: Sn are not collapse for n=0} and \ref{example: Sn are not collapse for n>0}.

To establish the agreement theorem for the second sequence, we first prove several preliminary lemmas.
\begin{lem}
\label{lem: less is better}
    Let $AF = \langle A, R \rangle$ be an argumentation framework, $S \subseteq A$ and $x,y\in A \setminus S$. Suppose that $x \sqsubseteq y$. Then $y \in \mathsf{u}(S)$ implies $ x \in \mathsf{u}(S)$.
\end{lem}
\begin{proof}
    If $y \in \mathsf{u}(S)$, then $y \in A \setminus (S \cup S^+)$, and hence $y \notin S^+$, which implies by $x \sqsubseteq y$ that $x \notin S^+$, it then follows from $x \in A \setminus S$ that $x \in \mathsf{u}(S)$.
\end{proof}

Recall that an argumentation framework is finitary iff each argument in it has only finitely many attackers.

\begin{lem}
    \label{lem: existence of minimal strict predecessor}
    Let $AF = \langle A, R \rangle$ be a finitary argumentation framework, $S \subseteq A$ and $x \in \mathsf{u}(S)$. 
    Suppose that $\vec{R}yx$ for some $y \in \mathsf{u}(S)$. 
    Then there exist some $z \in \mathsf{u}(S)$ such that $\vec{R}zx$, $z \sqsubseteq y$, and $z$ is $\sqsubset$-minimal in $x^-$. 
\end{lem}
\begin{proof} 
    Because $x \in \mathsf{u}(S)$, $x$ is not attacked by $S$. 
    Thus, $x^- \cap S = \textrm{Ø}$, which implies $x^- \subseteq A \setminus S$.
    Since $\vec{R}yx$, we have $y \in x^- \subseteq A \setminus S$.
    Because $AF$ is finitary, the set $x^-$ is finite, then there must exist some element $z \in x^- \subseteq A \setminus S$ such that $z$ is $\sqsubset$-minimal in $x^-$ and $z \sqsubseteq y$.
    Furthermore, we have by supposition $y \in \mathsf{u}(S)$. It then follows from Lemma \ref{lem: less is better} that $z \in \mathsf{u}(S)$.
    
    Next, we verify that $\vec{R}zx$.
    We already know $Rzx$ because $z \in x^-$.
    Additionally, since $z \sqsubseteq y$, we have $z^- \subseteq y^-$. 
    From our assumption $\vec{R}yx$, it follows that $x$ does not attack $y$, 
    Consequently, $x$ does not attack $z$. Therefore, $\vec{R}zx$ holds.
\end{proof}

\begin{lem}
\label{lem: minimal 2-prepending implies admissibly backward 2-extendable}
    Let $AF = \langle A, R \rangle$ be a finitary argumentation framework and $\mathcal{P}$ a set of paths. If $\mathcal{P}$ is closed under $\sqsubset$-minimal 2-prepending, then it is admissibly backward 2-extendable.
\end{lem}
\begin{proof}
    Let $S \subseteq A$ be an admissible set. To show that $\mathcal{P}$ is admissibly backward 2-extendable, we must demonstrate that $\mathcal{P}$ is strictly backward $2$-extendable with respect to $\mathsf{u}(S)$. 
    
    Let $\boldsymbol{p} \in \mathcal{P}{\downarrow}_{\mathsf{u}(S)}$ be a path originating at some $a \in \mathsf{u}(S)$, and 
    let $b \in \mathsf{u}(S)$ be a strict attacker of $a$ (i.e., $\vec{R}ba$).
     Assume there exists some $c \in \mathsf{u}(S)$ such that $\vec{R}cb$.
    By lemma \ref{lem: existence of minimal strict predecessor}, 
    there is some $c' \in \mathsf{u}(S)$ such that $\vec{R}c'b$ and $c'$ is $\sqsubset$-minimal in $b^-$.
    Finally, since $\mathcal{P}$ is closed under $\sqsubset$-minimal 2-prepending, the concatenated path $\langle c', b \rangle \circ \boldsymbol{p}$ must belong to $\mathcal{P}$; moreover, since $\boldsymbol{p}\in \mathcal{P}{\downarrow}_{\mathsf{u}(S)}$ and $b,c'\in\mathsf{u}(S)$, it also belongs to $\mathcal{P}{\downarrow}_{\mathsf{u}(S)}$. This completes the proof.
\end{proof}

\begin{lem}
\label{lem: existence of minimal strict path}
    Let $AF = \langle A, R \rangle$ be a finitary argumentation framework and $S \subseteq A$. Suppose that $\mathsf{u}(S)$ contains no $\vec{R}$-minimal elements. Then for all $n \geq 0$ and all $x \in \mathsf{u}(S)$, there is a $\sqsubset$-minimal strict path of length $n$ located entirely in $\mathsf{u}(S)$ that ends at $x$.
\end{lem}
\begin{proof}
We prove it by induction on $n$.

The base case $n=0$ is trivial, as the trivial path of length 0 from $x$ to $x$ is $\sqsubset$-minimal and located entirely in $\mathsf{u}(S)$.

For the inductive step. Let $x\in A$, by I.H., 
there is a $\sqsubset$-minimal strict path $\boldsymbol{p}$ of length $n$ located entirely in $\mathsf{u}(S)$ that ends at $x$. 
Assume that $\boldsymbol{p}$ originates at $a$.
 Since $\mathsf{u}(S)$ contains no $\vec{R}$-minimal elements, there is some $b \in \mathsf{u}(S)$ such that $\vec{R}ba$.
  By Lemma \ref{lem: existence of minimal strict predecessor}, there is some $c \in \mathsf{u}(S)$ such that $\vec{R}ca$ 
  and $c$ is $\sqsubset$-minimal in $a^-$. 
Then the concatenated path $\langle c \rangle \circ \boldsymbol{p}$ is a $\sqsubset$-minimal strict path of length $n+1$ located entirely in $\mathsf{u}(S)$ that ends at $x$.
\end{proof}

\begin{thm}
\label{thm: the agreement theorem by the second sequence}
    Let $AF=\left\langle A,R\right\rangle$ be a finitary argumentation framework that is free of self-defeating, and $n \in \mathbb{N}$.
    Suppose that $\prec^{\mathcal{S}_n}$ is well-founded. Then all preferred extensions of $AF$ are stable, and hence there is a stable extension of $AF$.
\end{thm}

\begin{proof}
    Since  $\mathcal{S}_n$ is closed under $\sqsubset$-minimal 2-prepending, by Lemma \ref{lem: minimal 2-prepending implies admissibly backward 2-extendable}, it is admissibly backward 2-extendable.

    We will verify the conditions of Theorem \ref{thm: the fundamental agreement theorem}. Let $S \subseteq A$ be an admissible set such that $\mathsf{u}(S)$ is nonempty. Because $\prec^{\mathcal{S}_n}$ is well-founded on $A$, $\prec^{\mathcal{S}_n}_{\mathsf{u}(S)}$ is also well-founded (by Lemma \ref{lem: monotonicity of path relation w.r.t. argument sets}), guaranteeing there is a $\prec^{\mathcal{S}_n}_{\mathsf{u}(S)}$-minimal element $x$ in $\mathsf{u}(S)$. If $\mathsf{u}(S)$ contains no $\vec{R}$-minimal element, then by Lemma \ref{lem: existence of minimal strict path}, there is a $\sqsubset$-minimal strict path of length $n$ located entirely in $\mathsf{u}(S)$ that ends at $x$, satisfying condition (2) of Theorem \ref{thm: the fundamental agreement theorem}. 

    Therefore, the conditions of Theorem \ref{thm: the fundamental agreement theorem} are met, allowing us to conclude that all preferred extensions of $AF$ are stable.
\end{proof}

\begin{cor}
    \label{cor:all minimal strict path has bounded length}
    Let $AF=\left\langle A,R\right\rangle$ be a finitary argumentation framework that is free of self-defeating, and $n \in \mathbb{N}$.
    Suppose that every $\sqsubset$-minimal strict path in $AF$ has length at most $n$. Then all preferred extensions of $AF$ are stable, and hence there is a stable extension of $AF$.
\end{cor}
\begin{proof}
    By supposition, $\mathcal{S}_{n+1}$ is empty, and hence $\prec^{\mathcal{S}_{n+1}}$ is well-founded, it then follows from Theorem \ref{thm: the agreement theorem by the second sequence} that all preferred extensions of $AF$ are stable.
\end{proof}

\section{Agreement Conditions in Finite Argumentations}
\label{Sect: Agreement Conditions in Finite Argumentations}
In this section, we investigate the conditions for agreement in finite argumentation frameworks, derive corollaries by combining these conditions with our agreement theorems, and provide examples that separate the well-foundedness conditions arising from the two sequences.

The following two lemmas generalize Lemma \ref{lem: Properties of relation based on E2 and O} and can be proved using a similar argument. 
The first lemma shows that both $\prec^{\mathcal{P}_{n}}$ and $\prec^{\mathcal{S}_{n}}$ are transitive whenever $n$ is even, and 
the second lemma shows that $\prec^{\mathcal{P}_{n}}$ satisfies a weak version of transitivity whenever $n$ is odd.

\begin{lem}
\label{lem: transitivity of the even relations}
    Let $AF=\left\langle A,R\right\rangle $ be an argumentation framework, and $\mathcal{P} =  \mathcal{P}_{2n}$ or $ \mathcal{P} = \mathcal{S}_{2n}$ for some $n\in\mathbb{N}$.
    For all $a,b,c \in A$, if $ a \prec^{\mathcal{P}} b \prec^{\mathcal{P}} c$, then $a \prec^{\mathcal{P}} c$.
\end{lem}
\begin{proof}
    Suppose $a \prec^{\mathcal{P}} b$ and $b \prec^{\mathcal{P}} c$. 
    By definition of $a \prec^{\mathcal{P}} b$, there exist paths $\boldsymbol{p}_{1}, \boldsymbol{p}_{1}' \in \mathcal{P}$ terminating at $b$, where $\boldsymbol{p}_{1}$ originates from $a$, and $\boldsymbol{p}_{1}'$ originates from some $b'$ such that $Rab'$. 
    Furthermore, the relation $b \prec^{\mathcal{P}} c$ yields a path $\boldsymbol{p}_{2} \in \mathcal{P}$ from $b$ to $c$. 
    The concatenation of $\boldsymbol{p}_{1}$ and $\boldsymbol{p}_{2}$ forms a path in $\mathcal{P}$ from $a$ to $c$. 
    By concatenating $\boldsymbol{p}_{1}'$ and $\boldsymbol{p}_{2}$, we obtain a path in $\mathcal{P}$ from $b'$ to $c$. 
    Since $Rab'$, the existence of these concatenated paths immediately establishes $a \prec^{\mathcal{P}} c$.
\end{proof}

\begin{lem}
\label{lem: weak transitivity of the odd relations}
    Let $AF=\left\langle A,R\right\rangle $ be an argumentation framework, and $\mathcal{P} =  \mathcal{P}_{2n+1}$ for some $n\in\mathbb{N}$. 
    For all $a,b,c,d \in A$, if $ a \prec^{\mathcal{P}} b \prec^{\mathcal{P}} c \prec^{\mathcal{P}} d$, then $a \prec^{\mathcal{P}} d$.
\end{lem}
\begin{proof}
    The proof closely mirrors that of Lemma \ref{lem: transitivity of the even relations}, except that it unpacks the chained three relations in $\mathcal{P}$ into successive odd paths, and then concatenate them to get another odd path in $\mathcal{P}$. 
\end{proof}

Note that the analogous weak transitivity property fails for 
$\prec^{\mathcal{S}_{2n+1}}$, since the concatenation of two paths in $\mathcal{S}_{2n+1}$ need not itself belong to $\mathcal{S}_{2n+1}$.

\begin{prop}
    \label{prop: the finite condition for the even relations}
    Let $AF=\left\langle A,R\right\rangle $ be a finite argumentation framework, and $\mathcal{P} =  \mathcal{P}_{2n}$ or $ \mathcal{P} = \mathcal{S}_{2n}$ for some $n\in\mathbb{N}$. 
    Then $\prec^{\mathcal{P}}$ is well-founded iff there is no $a \in A$ such that $a \prec^{\mathcal{P}} a$.
\end{prop}
\begin{proof}
    The left to the right direction is obvious. 
    Suppose that $\prec^{\mathcal{P}}$ is not well-founded.
    Then there exists an infinite descending $\prec^{\mathcal{P}}$-chain ${\langle a_{i} \rangle} _{i \in \mathbb{N}}$ such that $a_{i+1} \prec^{\mathcal{P}} a_{i}$ for each $i \in \mathbb{N}$. 
    Since $AF$ is finite, some $a_i$ occurs infinitely often in the chain, and hence $a_{i}=a_{j}$ for some $i<j \in \mathbb{N}$. 
    By Lemma \ref{lem: transitivity of the even relations}, $\prec^{\mathcal{P}}$ is transitive, and thus $a_{i} \prec^{\mathcal{P}} a_{j}$, which gives $a_i \prec^{\mathcal{P}} a_i$. 
\end{proof}

\begin{prop}
    \label{prop: the finite condition for the odd relations}
    Let $AF=\left\langle A,R\right\rangle $ be a finite argumentation framework, and $\mathcal{P} =  \mathcal{P}_{2n+1}$ for some $n\in\mathbb{N}$. 
    Then $\prec^{\mathcal{P}}$ is well-founded iff there do not exist $a,b \in A$ such that $a \prec^{\mathcal{P}} b$ and $b \prec^{\mathcal{P}} a$.
\end{prop}
\begin{proof}
    The left to the right direction is obvious. 
    Suppose that $\prec^{\mathcal{P}}$ is not well-founded.
    By the same argument as the proof of Proposition \ref{prop: the finite condition for the even relations}, there exists an infinite descending $\prec^{\mathcal{P}}$-chain ${\langle a_{i} \rangle} _{i \in \mathbb{N}}$ such that $a_{i} = a_{j}$ for some $i<j \in \mathbb{N}$. 
    If $j = i+1$, then $a_i \prec^{\mathcal{P}} a_i$; 
    if $j = i + 2$, then $a_i \prec^{\mathcal{P}} a_{i+1} \prec^{\mathcal{P}} a_j$, and thus $a_i \prec^{\mathcal{P}} a_{i+1}$ and $a_{i+1} \prec^{\mathcal{P}} a_i$; 
    otherwise, we have $j > i+2$, it then follows from Lemma \ref{lem: weak transitivity of the odd relations} that either $a_i \prec^{\mathcal{P}} a_j$ or $a_i \prec^{\mathcal{P}} a_{i+1} \prec^{\mathcal{P}} a_j$, and again we obtain that either $a_i \prec^{\mathcal{P}} a_i$, or $a_i \prec^{\mathcal{P}} a_{i+1}$ and $a_{i+1} \prec^{\mathcal{P}} a_i$.
\end{proof}

Recall that a path (or cycle) $a_0,\ldots,a_n$ traverses a symmetric edge if $\overset{\leftrightarrow}{R}a_ia_{i+1}$ for some $0\leqslant i\leqslant n-1$.
Since $a \prec^{\mathcal{P}_{2}} a$ requires existence of both a strict even cycle and an odd cycle traversing at most one symmetric edge,
the following two corollaries follows by Theorem \ref{thm: the agreement theorem by the first sequence} and Proposition \ref{prop: the finite condition for the even relations}.

\begin{cor}
    \label{cor: no strict cycle sufficent}
    Let $AF = \langle A,R \rangle$ be a finite argumentation framework that is free of self-defeating. 
    If $AF$ does not contain strict even cycles, then all preferred extensions of $AF$ are stable, and hence there is a stable extension of $AF$.
\end{cor}
\begin{proof}

    If $AF$ does not contain strict even cycles, then $a \prec^{\mathcal{P}_{2}} a$ for no $a \in A$, and hence we have by Proposition \ref{prop: the finite condition for the even relations} that $\prec^{\mathcal{P}_{2}}$ is well-founded, which implies by Theorem \ref{thm: the agreement theorem by the first sequence} that all preferred extensions of $AF$ are stable.
\end{proof}

The following corollary recovers Theorem 4.3 in \cite{duchet1980graphes}.
\begin{cor}
    \label{cor: no odd cycle with bidirected edge}
    Let $AF = \langle A,R \rangle$ be a finite argumentation framework. 
    If $AF$ does not contain odd cycles traversing at most one symmetric edge, then all preferred extensions of $AF$ are stable, and hence there is a stable extension of $AF$.
\end{cor}
\begin{proof}
    Assume that $AF$ does not contain odd cycles traversing at most one symmetric edge. Then $AF$ is free of self-defeating and $a \prec^{\mathcal{P}_{2}} a$ for no $a \in A$, which implies by Proposition \ref{prop: the finite condition for the odd relations} that $\prec^{\mathcal{P}_{2}}$ is well-founded, and hence by Theorem \ref{thm: the agreement theorem by the first sequence}, all preferred extensions of $AF$ are stable.
\end{proof}

\begin{example}
\label{example: p2 but not p0}
    Let $AF=\left\langle A,R\right\rangle$ be an argumentation framework depicted in Figure \ref{fig:triangle}, where $A=\{a,b,c\}$ and $R=\{\langle a,b \rangle,\langle b,a\rangle,\langle b,c\rangle,\langle c,a\rangle\}$. Since $a \prec^{\mathcal{P}_0} a$, we have that $\prec^{\mathcal{P}_0}$ is not well-founded. However, $\prec^{\mathcal{P}_2}$ is well-founded, otherwise it implies the presence of a strict even cycle with length  $> 0$, which is impossible. Therefore, $AF$ constitutes a counterexample of the converse of  (\ref{thm: collapse theorem: eum 3}) in Collapse Theorem. 

    \begin{figure}[htbp]
    \centering
    \begin{tikzpicture}[
        argument/.style={
            circle,
            draw,
            thick,
            minimum size=9mm,
            inner sep=0pt
        },
        attack/.style={
            -{Stealth[length=3mm,width=2.5mm]},
            thick
        }
    ]

    \node[argument] (a) at (0,1.2) {$a$};
    \node[argument] (b) at (2.5,1.2) {$b$};
    \node[argument] (c) at (1.25,0) {$c$};

    \draw[attack] (a) to[bend left=15] (b);
    \draw[attack] (b) to[bend right=15] (a);

    \draw[attack] (b) -- (c);
    \draw[attack] (c) -- (a);

    \end{tikzpicture}
\caption{}
\label{fig:triangle}
\end{figure}

\end{example}

The following concept of SCC-symmetric is introduced in \cite{baroni2007characterizing},
which is used to formulate a sufficient condition for the agreement
between stable and preferred semantics.

\begin{defn}
Let $AF=\left\langle A,R\right\rangle $ be an argumentation framework.
$AF$ is \emph{SCC-symmetric} iff for every attack cycle $a_{0},a_{1},\ldots,a_{n}$,
it holds that $\forall i\in\{1,...,n\},Ra_{i}a_{i-1}.$
\end{defn}

The following statement has been proved in \cite{baroni2007characterizing},
see Corollary $1$.
\begin{cor}
Let $AF=\left\langle A,R\right\rangle $ be a finite argumentation
framework that is free of self-defeating. If $AF$ is SCC-symmetric,
then all preferred extensions in $AF$ are stable, and hence there is a stable extension of $AF$.
\end{cor}

\begin{proof}
Since $AF$ is free of self-defeating, the length of each odd cycle
in $AF$ is greater than $1$, it then follows from the SCC-symmetry
of $AF$ that it traverses more than one symmetric edge. Hence,
there is no odd cycle in $AF$ traversing at most one symmetric edge, which
implies by Corollary \ref{cor: no odd cycle with bidirected edge} that all preferred
extensions in $AF$ are stable.
\end{proof}

The following corollary reformulates the well-foundedness condition for $\prec^{\mathcal{P}_1}$ in terms of strict odd paths and cycles.
\begin{cor}
    \label{cor: no overlap cycles for P_1}
    Let $AF = \langle A,R \rangle$ be a finite argumentation framework that is free of self-defeating. 
Suppose that $AF$ contains no strict odd cycle and that there do not exist distinct arguments $a,b$ such that:
\begin{enumerate}
    \item there are strict odd paths from $a$ to $b$ and from $c$ to $b$, with $Rac$ for some $c$; and
    \item there are strict odd paths from $b$ to $a$ and from $d$ to $a$, with $Rbd$ for some $d$.
\end{enumerate}
    Then all preferred extensions of $AF$ are stable, and hence there is a stable extension of $AF$.
\end{cor}
\begin{proof}
    By supposition, there are neither $x \in A$ with $x \prec^{\mathcal{P}_1} x$ nor distinct $a,b \in A$ such that $a \prec^{\mathcal{P}_1} b$ and $b \prec^{\mathcal{P}_1} a$. It then follows from Proposition \ref{prop: the finite condition for the odd relations} that $\prec^{\mathcal{P}_1}$ is well-founded, and hence by Theorem \ref{thm: the agreement theorem by the first sequence}, all preferred extensions of $AF$ are stable.
\end{proof}

\begin{example}
\label{example: p1 but not p2}
Let $AF=\left\langle A,R\right\rangle$ be the argumentation
framework depicted in Figure~\ref{fig:two-cycles}, where
$A=\{a,b,c,d,e,f\}$ and
$R=\{
    \langle a,b\rangle,
    \langle b,c\rangle,
    \langle c,d\rangle,
    \langle d,a\rangle,
    \langle a,e\rangle,
    \langle e,a\rangle,
    \langle e,f\rangle,
    \langle f,a\rangle
    \}.$
Since $a \prec^{\mathcal{P}_2} a$, the relation $\prec^{\mathcal{P}_2}$ is not well-founded. It is straightforward to verify that $AF$ contains no strict odd cycle. Moreover, no pair of distinct arguments satisfies both conditions (1) and (2) of Corollary \ref{cor: no overlap cycles for P_1}. Indeed, for each $x \neq a,e$ and $y$, it is not the case that $x \prec^{\mathcal{P}_1} y$, because whenever there is a strict odd path from $x$ to $y$, every strict path from an argument $z$ with $Rxz$ to $y$ has even length. Furthermore, neither $a \prec^{\mathcal{P}_1} e$ nor $e \prec^{\mathcal{P}_1} a$. Hence, $\prec^{\mathcal{P}_1}$ is well-founded. Therefore, $AF$ provides a counterexample to the converse of (\ref{thm: collapse theorem: eum 4}) in the Collapse Theorem.

\begin{figure}[htbp]
    \centering
    \begin{tikzpicture}[
        argument/.style={
            circle,
            draw,
            thick,
            minimum size=9mm,
            inner sep=0pt
        },
        attack/.style={
            -{Stealth[length=3mm,width=2.5mm]},
            thick
        }
    ]

        \node[argument] (c) at (0,3.6)    {$c$};
        \node[argument] (d) at (-1.5,2.4) {$d$};
        \node[argument] (b) at (1.5,2.4)  {$b$};
        \node[argument] (a) at (0,1.5)    {$a$};
        \node[argument] (f) at (-1.5,0)   {$f$};
        \node[argument] (e) at (1.5,0)    {$e$};

        \draw[attack] (a) -- (b);
        \draw[attack] (b) -- (c);
        \draw[attack] (c) -- (d);
        \draw[attack] (d) -- (a);

        \draw[attack] (a) -- (e);
        \draw[attack] (e) -- (a);
        \draw[attack] (e) -- (f);
        \draw[attack] (f) -- (a);

    \end{tikzpicture}
    \caption{}
    \label{fig:two-cycles}
\end{figure}

\end{example}

\begin{cor}
    \label{cor: no cycle in S_2n is sufficent for agreement}
    Let $AF = \langle A,R \rangle$ be a finite argumentation framework that is free of self-defeating and $n \in \mathbb{N}$. 
    If ${\mathcal{S}_{2n}}$ does not contain cycles, then all preferred extensions of $AF$ are stable, and hence there is a stable extension of $AF$.
\end{cor}
\begin{proof}
    Assume a preferred extension is not stable. By Theorem \ref{thm: the agreement theorem by the second sequence} and Proposition \ref{prop: the finite condition for the even relations}, there is some $a \in A$ such that $a \prec^{\mathcal{S}_{2n}} a$, which implies that a path in $\mathcal{S}_{2n}$ forms a cycle.
\end{proof}

\begin{cor}
    Let $AF = \langle A,R \rangle$ be a finite argumentation framework that is free of self-defeating and $n \in \mathbb{N}$. 
    If there does not exist $\langle a_0, \dots, a_k \rangle \in \mathcal{S}_{2n}$ with $Ra_ka_0$, then all preferred extensions of $AF$ are stable, and hence there is a stable extension of $AF$.
\end{cor}
\begin{proof}
    If $AF$ has a non-stable preferred extension, then by Theorem \ref{thm: the agreement theorem by the second sequence} and Proposition \ref{prop: the finite condition for the even relations}, $a \prec^{\mathcal{S}_{2n}} a$ for some $a \in A$, which dictates that there exists a path $\langle a_0, \dots, a_k \rangle \in \mathcal{S}_{2n}$ with $Ra_k a_0$.
\end{proof}

\begin{cor}
    \label{cor:no minimal cycle}
    Let $AF = \langle A,R \rangle$ be a finite argumentation framework that is free of self-defeating. 
    If there does not exist any strict $\sqsubset$-minimal cycle, then all preferred extensions of $AF$ are stable, and hence there is a stable extension of $AF$.
\end{cor}
\begin{proof}
If there does not exist any strict $\sqsubset$-minimal cycle, then every strict $\sqsubset$-minimal path has length at most $n$, where $n$ is the number of arguments in $A$. It then follows from Corollary \ref{cor:all minimal strict path has bounded length} that all preferred extensions of $AF$ are stable.
\end{proof}

The following two examples demonstrate that the well-foundedness of $\prec^{\mathcal{S}_n}$ and that of $\prec^{\mathcal{S}_m}$ need not coincide when $m>n$. The first example treats the case $m>n=0$, while the second treats the case $m>n>0$.

\begin{example}
\label{example: Sn are not collapse for n=0}
Let $AF=\left\langle A,R\right\rangle$ be the argumentation
framework depicted in Figure~\ref{fig:triangle with tail}, where
$
A=\{a,b,c,d\}
$
and
$
R=
\{
\langle a,b\rangle,
\langle d,b\rangle,
\langle b,c\rangle,
\langle c,d\rangle
\}.
$

Since $c$ is $\sqsubset$-minimal in $d^-$, we obtain
$$
\langle c,d,b\rangle\in\mathcal{S}_0.
$$
Together with $\langle b\rangle\in\mathcal{S}_0$ and $Rbc$, this witnesses $b\prec^{\mathcal{S}_0}b$. Hence, $\prec^{\mathcal{S}_0}$ is not well-founded.

Now let $m>0$. First, observe that $a$ is $\sqsubset$-minimal in $b^-$, whereas $d$ is not. The failure of the $\sqsubset$-minimality requirement occurs when $b$ is immediately preceded by $d$. The longest $\sqsubset$-minimal strict path is $\langle a,b,c,d\rangle$; hence, $\mathcal{S}_m=\varnothing$ for $m\geq4$, while $\mathcal{S}_3$ contains only this path. Hence, $\prec^{\mathcal{S}_m}$ is empty whenever $n \geq 3$.

For $m=2$, any cycle in $\mathcal{S}_2$ would have to traverse $\langle b,c,d,b\rangle$ at least twice. Since successive occurrences of $d$ are three positions apart, one occurrence would violate either the $\sqsubset$-minimality of the terminal segment or the closure condition under $\sqsubset$-minimal $2$-prepending. Thus, $\mathcal{S}_2$ contains no cycles, which implies by Proposition \ref{prop: the finite condition for the even relations} that $\prec^{\mathcal{S}_2}$ is well-founded.

For $m=1$, since $d$ is not $\sqsubset$-minimal in $b^-$, no path in $\mathcal{S}_1$ can originate at $d$. Consequently, neither $d$ nor $c$ can precede another argument under $\prec^{\mathcal{S}_1}$: the former originates no path, while the latter attacks only $d$, so $c\prec^{\mathcal{S}_1}x$ would require a path originating at $d$.
On the other hand, no path in $\mathcal{S}_1$ terminates at $a$, and $\langle a,b\rangle$ is the only path terminating at $b$. Thus, there is no \(x\) such that \(x \prec^{\mathcal S_1} a\) or \(x \prec^{\mathcal S_1} b\). It follows that every possible $\prec^{\mathcal{S}_1}$-chain must begin at $a$ or $b$ and terminate at $c$ or $d$, where it cannot continue. Hence, $\prec^{\mathcal{S}_1}$ is well-founded.

\begin{figure}[htbp]
    \centering
    \begin{tikzpicture}[
        argument/.style={
            circle,
            draw,
            thick,
            minimum size=9mm,
            inner sep=0pt
        },
        attack/.style={
            -{Stealth[length=3mm,width=2.5mm]},
            thick
        }
    ]

        \node[argument] (a) at (-2.5,0)  {$a$};
        \node[argument] (b) at (0,0)     {$b$};
        \node[argument] (c) at (1.25,2.165) {$c$};
        \node[argument] (d) at (2.5,0)   {$d$};

        \draw[attack] (a) -- (b);
        \draw[attack] (b) -- (c);
        \draw[attack] (c) -- (d);
        \draw[attack] (d) -- (b);

    \end{tikzpicture}

    \caption{}
    \label{fig:triangle with tail}
\end{figure}
\end{example}

\begin{example}
\label{example: Sn are not collapse for n>0}
Let
$AF_n=\left\langle A_n,R_n\right\rangle$ be the argumentation
framework with $n \geq 1$ depicted in Figure~\ref{fig:Sn are not collapse for n>0}, where
\[
A_n
=
\{z,z',x\}
\cup
\{y_i\mid 1\leq i\leq n+1\}
\cup
\{x_i\mid 0\leq i\leq n+1\},
\]
and
\[
\begin{aligned}
R_n={}&
\bigl\{
    \langle z',z\rangle,
    \langle z,y_2\rangle,
    \langle z',y_1\rangle,
    \langle y_1,z'\rangle,
    \langle x,y_1\rangle,\\
&\qquad
    \langle y_{n+1},x\rangle,
    \langle x,x_0\rangle,
    \langle x_0,x\rangle,
    \langle x_{n+1},x\rangle
\bigr\}\\
&\cup
\bigl\{
    \langle y_i,y_{i+1}\rangle
    \mid 1\leq i\leq n
\bigr\}\\
&\cup
\bigl\{
    \langle x_i,x_{i+1}\rangle
    \mid 0\leq i\leq n
\bigr\}.
\end{aligned}
\]
Consider the cycle
\[
    C=\langle x,y_1,y_2,\ldots,y_{n+1},x\rangle .
\]
The suffix $\langle y_2,\ldots,y_{n+1},x\rangle$ is a \(\sqsubset\)-minimal path of length \(n\). 
Moreover, \(x\) is a \(\sqsubset\)-minimal attacker of \(y_1\). Since $\mathcal{S}_n$ is closed under
\(\sqsubset\)-minimal \(2\)-prepending,  $C\in\mathcal{S}_n$.
The path
\[
    \langle x_0,x_1,\ldots,x_{n+1},x\rangle
\]
is a \(\sqsubset\)-minimal path of length \(n+2\), and also belongs to \(\mathcal{S}_n\). It then follows from \(x\) attacks \(x_0\) that  $x\prec^{\mathcal{S}_n}x$.
Therefore, \(\prec^{\mathcal{S}_n}\) is not well-founded.

Let $m>n$. We first show that $\mathcal{S}_m$ does not contain strict cycle. Observe that every strict cycle $C'$ is obtained by choosing a starting point on $C$ and traversing $C$ one or more times, and that every path in $\mathcal{S}_m$ has a $\sqsubset$-minimal terminal segment of length $m$. If $m\geq n+2$, the terminal segment of $C'$ contains a complete traversal of $C$, and hence contains consecutive arguments $y_1,y_2$. Since $y_1$ is not $\sqsubset$-minimal in $y_2^-$, this terminal segment cannot be $\sqsubset$-minimal. 

It remains to consider $m=n+1$. If $n$ is even, then $C$ has even length, thus every strict cycle has even length, whereas every path in $\mathcal{S}_m$ has odd length. If $n$ is odd, then a single traversal of $C$ has odd length and hence cannot belong to $\mathcal{S}_m$. Any repeated traversal contains at least two occurrences of $y_1,y_2$. 
Since the length of $C$ is odd, their positions have alternating parity. Thus, some occurrence of $y_1$ violates the $\sqsubset$-minimal condition. Hence, no strict cycle belongs to $\mathcal{S}_m$.

Suppose, for the sake of contradiction, that $\prec^{\mathcal{S}_m}$ were not well-founded. Then there would be an infinite descending $\mathcal{S}_m$-chain. First, note that $\{z',x_0\}$ can not occur in the descending $\prec^{\mathcal{S}_m}$-chain, because no path in $\mathcal{S}_m$ can terminate at them. Second, there are only two $\sqsubset$-minimal strict paths of length at least $n+1$ terminating at arguments outside $\{x,x_0,y_1,z'\}$:
$$
\langle z',z,y_2,\ldots,y_{n+1}\rangle
\qquad\text{and}\qquad
\langle x_0,x_1,\ldots,x_{n+1}\rangle.
$$
However, these two paths cannot produce $\prec^{\mathcal{S}_m}$-pairs in the descending $\prec^{\mathcal{S}_m}$-chain. All other paths terminating outside $\{x,x_0,y_1,z'\}$ are either too short or have a terminal segment containing $y_1,y_2$, and hence do not belong to $\mathcal{S}_m$. Therefore, the only arguments that can occur repeatedly in the descending chain are $x$ and $y_1$.

Consequently, the acyclicity of $\mathcal{S}_m$ implies that the descending $\mathcal{S}_m$-chain can only be formed by 
\[
    x\prec^{\mathcal{S}_m}y_1
    \qquad\text{and}\qquad
    y_1\prec^{\mathcal{S}_m}x.
\]
This would require a path in $\mathcal{S}_m$ from $y_1$ to $x$. Every such path ends with
$$
\langle y_1,y_2,\ldots,y_{n+1},x\rangle,
$$
which has length $n+1$. It must therefore be the $\sqsubset$-minimal terminal segment of the path. This is impossible because $y_1$ is not $\sqsubset$-minimal in $y_2^-$. Hence, no infinite descending chain exists, and $\prec^{\mathcal{S}_m}$ is well-founded.

Consequently, the well-foundedness of
\(\prec^{\mathcal{S}_m}\) and that of \(\prec^{\mathcal{S}_n}\) are
not equivalent whenever \(m>n\geq 1\).

\begin{figure}[htbp]
    \centering
    \begin{tikzpicture}[
        argument/.style={
            circle,
            draw,
            thick,
            minimum size=9mm,
            inner sep=0pt
        },
        attack/.style={
            -{Stealth[length=3mm,width=2.5mm]},
            thick
        }
    ]

        \node[argument] (zp) at (-5,3.6) {$z'$};
        \node[argument] (z)  at (-5,0)   {$z$};
        \node[argument] (y1) at (-3,3.6) {$y_1$};
        \node[argument] (y2) at (-3,0)   {$y_2$};

        \node[argument] (x)  at (0,3.6) {$x$};
        \node[argument] (x2) at (0,0)   {$x_2$};

        \node[argument] (x0) at (2,3.6) {$x_0$};
        \node[argument] (x1) at (2,0)   {$x_1$};

        \node[argument] (yn) at (-1.25,2.10) {$y_{n+1}$};
        \node[argument] (xn) at (0,2.1)    {$x_{n+1}$};

        \coordinate (ya) at (-2.37,0.756);
        \coordinate (yb) at (-1.88,1.344);
        \coordinate (xa) at (0,0.80);
        \coordinate (xb) at (0,1.30);

        \draw[attack] (zp) -- (y1);
        \draw[attack] (y1) -- (zp);

        \draw[attack] (zp) -- (z);
        \draw[attack] (z)  -- (y2);
        \draw[attack] (y1) -- (y2);

        \draw[attack] (x) -- (y1);

        \draw[attack] (y2) -- (ya);
        \node[rotate=50.2,inner sep=0pt] at (-2.125,1.05) {$\cdots$};
        \draw[attack] (yb) -- (yn);
        \draw[attack] (yn) -- (x);

        \draw[attack] (x)  -- (x0);
        \draw[attack] (x0) -- (x);

        \draw[attack] (x0) -- (x1);
        \draw[attack] (x1) -- (x2);

        \draw[attack] (x2) -- (xa);
        \node at (0,1.15) {$\vdots$};
        \draw[attack] (xb) -- (xn);
        \draw[attack] (xn) -- (x);

    \end{tikzpicture}

    \caption{}
    \label{fig:Sn are not collapse for n>0}
\end{figure}
\end{example}

The examples above also show that $\prec^{\mathcal{S}_n}$ may remain well-founded even when the framework contains strict odd cycles. Consequently, our agreement theorem of the second sequence applies beyond frameworks satisfying conventional odd-cycle-freeness conditions. One simple approach, as demonstrated in Corollary \ref{cor:no minimal cycle}, is to check whether strict cycles are $\sqsubset$-minimal.


\section{From Global to Local Semantic Agreement}
\label{Sect: From Global to Local Semantic Agreement}
This section establishes a reduction principle for semantic agreement that localizes our global conditions relative to an admissible set and yields strengthened local counterparts. Specifically, to guarantee that every preferred extension containing an admissible set \(S\) is stable, it suffices to verify the corresponding condition on the arguments that remain undecided relative to \(S\).

\begin{lem}
\label{lem: presistence ofextensions under restrictions}
Let $AF=\left\langle A,R\right\rangle$ be an argumentation framework, $S$ an admissible set in $AF$ and $E = S \cup E'$ for some $E' \subseteq \mathsf{u}(S)$.
Then we have:
\begin{enumerate}
    \item 
    \label{enu:lem: presistence ofextensions under restrictions 1}
    $E$ is admissible in $AF$ iff $E'$ is admissible in $AF{\downarrow}_{\mathsf{u}(S)}$;
    \item 
    \label{enu:lem: presistence ofextensions under restrictions 2}
    $E$ is a preferred extension of $AF$ iff $E'$ is a preferred extension of $AF{\downarrow}_{\mathsf{u}(S)}$;
    \item 
    \label{enu:lem: presistence ofextensions under restrictions 3}
    $E$ is a stable extension of $AF$ iff $E'$ is a stable extension of $AF{\downarrow}_{\mathsf{u}(S)}$.
\end{enumerate}
\end{lem}

\begin{proof}
(\ref{enu:lem: presistence ofextensions under restrictions 1}) Assume that $E$ is admissible in $AF$. Then $E$ is conflict-free in $AF$ and $E \subseteq \mathcal{F}_{AF}(E)$, 
and hence $E'$ is conflict-free in $AF{\downarrow}_{\mathsf{u}(S)}$. 
To show that $E' \subseteq\mathcal{F}_{AF{\downarrow}_{\mathsf{u}(S)}}(E')$,
let $a\in E'$ and $b\in\mathsf{u}(S)$ such that $Rba$. 
Because $E\subseteq\mathcal{F}_{AF}(E)$, we have that $Rcb$
for some $c\in E$. 
Since $b\in\mathsf{u}(S)$, we know $b\notin S^{+}$,
which implies that $c\notin S$, and thus $c\in E'$.
For the converse direction, assume that $E'$ is admissible in $AF{\downarrow}_{\mathsf{u}(S)}$. 
Then we have by Lemma \ref{lem: union of two ad-ext}
that $E = S \cup E'$ is admissible in $AF$.

(\ref{enu:lem: presistence ofextensions under restrictions 2}) Assume that $E$ is a preferred extension of $AF$. 
Then $E$ is a maximal admissible set in $AF$, 
which implies by (\ref{enu:lem: presistence ofextensions under restrictions 1}) that $E'$ is an admissible set in $AF{\downarrow}_{\mathsf{u}(S)}$.
Suppose that $E'$ is not a maximal admissible set in $AF{\downarrow}_{\mathsf{u}(S)}$. 
Then there is a nonempty set $S'\subseteq\mathsf{u}(S)-E'$ such that $E' \cup S'$ is an admissible set in $AF{\downarrow}_{\mathsf{u}(S)}$, 
which implies by Lemma \ref{lem: union of two ad-ext} that $E\cup S'=S \cup E'\cup S'$ is an admissible set in $AF$, 
contradicting the maximality of $E$. 
Hence, $E'$ is a maximal admissible set in $AF{\downarrow}_{\mathsf{u}(S)}$, and therefore is a preferred extension of $AF{\downarrow}_{\mathsf{u}(S)}$.

Assume that $E'$ is a preferred extension of $AF{\downarrow}_{\mathsf{u}(S)}$.
Then $E'$ is a maximal admissible set in $AF{\downarrow}_{\mathsf{u}(S)}$,
which implies by (\ref{enu:lem: presistence ofextensions under restrictions 1})
that $E$ is an admissible set in $AF$. 
Suppose that $E$ is not a maximal admissible set in $AF$. 
Then there is a nonempty $S'\subseteq A - E$ such that $S \cup E' \cup S'$ is an admissible set in $AF$, 
and thus it is conflict-free, which implies $S'\subseteq {\mathsf{u}(S)}$.
Hence, we have by (\ref{enu:lem: presistence ofextensions under restrictions 1}) that $E' \cup S'$ is admissible in $AF{\downarrow}_{\mathsf{u}(S)}$,
which contradicts to the maximality of $E'$. 
Therefore, $E$ is a maximal admissible set in $AF$, and hence is a preferred extension of $AF$.

(\ref{enu:lem: presistence ofextensions under restrictions 3}) Assume that $E$ is a stable extension of $AF$. 
Then $E=A-E^{+}$, and thus $E$ is conflict-free, 
which implies that $E' \subseteq \mathsf{u}(S)-(E')^{+}$.
To show that $\mathsf{u}(S)-(E')^{+} \subseteq E'$,
let $a\in\mathsf{u}(S)-(E')^{+}$. 
Since $\mathsf{u}(S)\cap S^{+}=\textrm{Ø}$, $a \notin (E'\cup S)^+ = E^+$.
It then follows from $E=A-E^{+}$ that $a \in E$,
and hence we obtain by $a \in \mathsf{u}(S)$ that $a \in E'$.
Therefore, $\mathsf{u}(S)-(E')^{+}\subseteq E'$.

Assume that $E'$ is a stable extension of $AF{\downarrow}_{\mathsf{u}(S)}$.
Then $E'=\mathsf{u}(S)-(E')^{+}$,
and thus $E'$ is conflict-free. Since $S$ is admissible
in $AF$ and does not attack any argument in $\mathsf{u}(S)$,
we know from Lemma \ref{fact: no attack each other} that $\mathsf{u}(S)$
does not attack any argument in $S$, which gives by $E' \subseteq \mathsf{u}(S)$
that $E$ is conflict-free, and
thus $E\subseteq A-E^{+}$. To show that $A-E^{+}\subseteq E$, let
$a\in A-E^{+}$. Since $S\subseteq E$, either $a\in S$ or $a\in\mathsf{u}(S)$.
If $a\in S$, then we have by $S\subseteq E$ that $a\in E$; if $a\in\mathsf{u}(S)$,
then we have by $a\in A-E^{+}$ that $a\in\mathsf{u}(S)-E^{+}\subseteq\mathsf{u}(S)-(E')^{+} = E' \subseteq E$. Therefore, $A-E^{+}\subseteq E$.
\end{proof}

The following proposition establishes the reduction principle underlying our local agreement results.
\begin{prop}
\label{prop: reduce agreement to the restricts}
Let $AF=\left\langle A,R\right\rangle $
be an argumentation framework and $S$ an admissible set in $AF$.
Suppose that all preferred extensions of $AF{\downarrow}_{\mathsf{u}(S)}$
are stable. Then all preferred extensions of $AF$ including $S$
are stable, and hence there is a stable extension of $AF$ including $S$.
\end{prop}

\begin{proof}
Let $E$ be any preferred extensions of $AF$ including $S$. 
Then $E = S \cup E'$ for some $E' \subseteq {\mathsf{u}(S)}$, 
and hence by Lemma \ref{lem: presistence ofextensions under restrictions}(\ref{enu:lem: presistence ofextensions under restrictions 2}),
$E'$ is a preferred extension of $AF{\downarrow}_{\mathsf{u}(S)}$,
it then follows from the supposition that $E'$ is a stable extension of $AF{\downarrow}_{\mathsf{u}(S)}$, 
which implies by Lemma \ref{lem: presistence ofextensions under restrictions}(\ref{enu:lem: presistence ofextensions under restrictions 3}) that $E$ is a stable extension of $AF$.
\end{proof}

\begin{cor}
\label{Cor: reduce agreement to the restrict induced by GE}
Let $AF=\left\langle A,R\right\rangle $
be an argumentation framework and $S$ the grounded (or ideal) extension
of $AF$. If all preferred extensions of $AF{\downarrow}_{\mathsf{u}(S)}$
are stable, then all preferred extensions of $AF$ are stable, and hence there is a stable extension of $AF$.
\end{cor}

\begin{proof}
If all preferred extensions of $AF{\downarrow}_{\mathsf{u}(S)}$ are
stable, then by Proposition \ref{prop: reduce agreement to the restricts},
all preferred extensions of $AF$ including $S$ are stable, since
$S$ is the grounded (or ideal) extension of $AF$, $S$ is a subset
of all preferred extensions of $AF$, and hence all preferred extensions
of $AF$ are stable.
\end{proof}

By localizing the well-foundedness condition to the set of undecided arguments, we obtain the following strengthened versions of Theorem \ref{thm: the agreement theorem by the first sequence} and \ref{thm: the agreement theorem by the second sequence}. 
This formulation is strictly stronger because it only requires well-foundedness on the restricted subframework $AF{\downarrow}_{\mathsf{u}(S)}$, allowing the global framework to violate the condition as long as the problematic arguments are resolved by $S$. Furthermore, taking $S = \emptyset$ immediately recovers the corresponding global agreement theorems as a special case. 

\begin{thm}
    Let $AF=\left\langle A,R\right\rangle$ be an argumentation framework, $S$ an admissible set in $AF$ attacking every self-defeating arguments, and $n \in \mathbb{N}$.
    Suppose that $\prec^{\mathcal{P}_n}_{\mathsf{u}(S)}$ is well-founded. Then all preferred extensions of $AF$ including $S$ are stable, and hence there is a stable extension of $AF$ including $S$.
\end{thm}
\begin{proof}
    By hypothesis, the relation $\prec^{\mathcal{P}}_{\mathsf{u}(S)}$ is well-founded. Applying Theorem \ref{thm: the agreement theorem by the first sequence} to this subframework, we deduce that all preferred extensions of $AF{\downarrow}_{\mathsf{u}(S)}$ are stable. Consequently, Proposition \ref{prop: reduce agreement to the restricts} guarantees that every preferred extension of $AF$ including the admissible set $S$ is stable.
\end{proof}

\begin{thm}
    Let $AF=\left\langle A,R\right\rangle$ be a finitary argumentation framework, $S$ an admissible set in $AF$ attacking every self-defeating arguments, and $n \in \mathbb{N}$.
    Suppose that $\prec^{\mathcal{S}_n}_{\mathsf{u}(S)}$ is well-founded. Then all preferred extensions of $AF$ including $S$ are stable, and hence there is a stable extension of $AF$ including $S$.
\end{thm}
\begin{proof}
    This can be proved similarly as the theorem above.
\end{proof}

All agreement corollaries can likewise be strengthened to their corresponding local versions. We omit these stronger formulations and proofs, as, like the results above, they follow immediately from their global counterparts together with Proposition \ref{prop: reduce agreement to the restricts}.

We say that $S$ \emph{attacks }a cycle if $S$ attacks some argument from the cycle. 
Recently, it has been shown in \cite{baumann2021cycles} that a stable extension exists in a finite abstract argumentation framework if and only if there exists an admissible set in it attacking every odd cycle. 
The following corollary generalizes the statement.

\begin{cor}
    \label{cor: a sufficent and necessary using odd cycles with at most one bidirected edge}
Let $AF=\langle A,R\rangle $ be a finite argumentation framework. 
There is a stable extension of $AF$ if and only if 
there exists an admissible set attacking every odd cycle traversing at most one symmetric edge.
\end{cor}

\begin{proof}
The left to the right direction is trivial, 
just let any stable extension be the required admissible set, 
which attacks every cycle in $AF$. 
To show the converse direction, assume that an admissible set $S$ attacks every
odd cycle traversing at most one symmetric edge. 
Then $AF{\downarrow}_{\mathsf{u}(S)}$ contains no odd cycle traversing at most one symmetric edge, 
which gives by Corollary \ref{cor: no odd cycle with bidirected edge} that all preferred
extensions of $AF{\downarrow}_{\mathsf{u}(S)}$ are stable.
It then follows from Proposition \ref{prop: reduce agreement to the restricts} that all preferred
extensions of $AF$ including $S$ are stable. 
By Theorem \ref{thm: preferred contain admissible set}, there is a preferred
extension $E$ in $AF$ such that $S\subseteq E$, 
and hence $E$ is a stable extension. 
\end{proof}

Additional necessary and sufficient conditions for the existence of stable extensions can be obtained by applying other agreement corollaries. In what follows, we use Corollaries \ref{cor: no strict cycle sufficent} and \ref{cor:no minimal cycle} as illustrative examples. The next two corollaries are analogous to the preceding result and can be proved by applying Corollaries \ref{cor: no strict cycle sufficent} and \ref{cor:no minimal cycle}, respectively, with essentially the same argument.

\begin{cor}
Let $AF=\langle A,R\rangle $ be a finite argumentation framework. 
There is a stable extension of $AF$ if and only if 
there is an admissible set attacking every self-defeating argument and every strict even cycle.
\end{cor}

\begin{cor}
\label{cor:necessary and sufficent condition of stable by minimal cycles}
Let $AF=\langle A,R\rangle $ be a finite argumentation framework. 
There is a stable extension of $AF$ if and only if 
there exists an admissible set $S$ such that $AF{\downarrow}_{\mathsf{u}(S)}$ contains neither self-defeating arguments nor strict $\sqsubset_{AF{\downarrow}_{\mathsf{u}(S)}}$-minimal cycles.
\end{cor}

\section{Conclusion}
\label{Sect: Conclusion}
In this paper, we have developed a general path-based framework for investigating when preferred and stable semantics agree in abstract argumentation frameworks. The framework is built around suitably chosen sets of paths and the relations induced by them, with the Fundamental Theorem providing the abstract basis for deriving agreement results. By instantiating this framework with different classes of paths, we obtained a range of concrete sufficient conditions expressed in terms of well-foundedness. In particular, the classical agreement result for limited controversial frameworks is recovered through the set of even paths. We further introduced two infinite sequences of path sets, each giving rise to a family of sufficient conditions for semantic agreement. The two families exhibit essentially different behaviour: the conditions associated with the first sequence collapse to three distinct well-foundedness conditions, whereas those associated with the second sequence remain genuinely distinct. The framework applies to both finite and infinite argumentation frameworks and provides a common setting in which these different agreement conditions can be studied. We also established a reduction principle for semantic agreement, allowing our
global agreement conditions to be localized relative to an admissible set. 

Several directions for further work arise naturally from this framework. The most immediate is to identify further path sets to which the Fundamental Theorem can be effectively applied. The path sets considered in this paper represent only particular ways of encoding structural information about attacks, and other choices may lead to new sufficient conditions that are independent of, or weaker than, those obtained here. A more systematic study of which path constructions yield useful induced relations, and of the relationships among the resulting conditions, may also help clarify the scope of the path-based method.

A further question is whether the path-based ideas developed here can be adapted to obtain more explicit structural conditions for the existence of stable extensions. The agreement conditions established in this paper guarantee stable extension existence by imposing the stronger requirement that every preferred extension be stable. We also obtain necessary and sufficient characterizations of stable extension existence in terms of the existence of an admissible set satisfying certain attack conditions. This characterization, however, still quantifies over admissible sets. It would therefore be interesting to identify weaker conditions stated directly in terms of the structure of the attack graph---for example, in terms of paths, cycles, or relations induced by suitable path sets---that guarantee the existence of at least one stable extension without requiring full agreement between preferred and stable semantics.

Finally, the relationship between abstract and assumption-based argumentation suggests a broader domain of application. Assumption-based argumentation can be instantiated as abstract argumentation, while abstract argumentation can in turn be represented within assumption-based argumentation; see, e.g., \cite{toni2014tutorial,cyras2018assumption}. An interesting question is therefore whether the conditions developed here admit natural formulations directly at the level of assumptions, rules, and contraries, rather than only at the level of abstract argumentation frameworks. Such formulations could make the results more informative for structured argumentation and may also reveal which features of the underlying system give rise to the path conditions studied in this paper. More broadly, given the well-established connections between argumentation semantics and semantics for logic programming and default reasoning, it would be worthwhile to investigate how the present path-based conditions translate into corresponding structural or syntactic conditions in these and other nonmonotonic formalisms.


\bibliographystyle{plain}
\bibliography{zy}
\end{document}